\documentclass{article}

\usepackage[utf8]{inputenc} 
\usepackage[T1]{fontenc}    
\usepackage{hyperref}       
\usepackage{url}            
\usepackage{booktabs}       
\usepackage{amsfonts}       
\usepackage{nicefrac}       
\usepackage{microtype}      
\usepackage{xcolor}         

\usepackage{graphicx}
\usepackage{doi}
\usepackage{amsmath, amsthm, amssymb} 
\usepackage{bbold}
\usepackage{caption}
\usepackage{subcaption}
\usepackage{multirow}
\usepackage{bm}
\usepackage{setspace}
\usepackage{comment}

\usepackage[preprint]{neurips_2022}

\title{Going Beyond One-Hot Encoding in Classification: Can Human Uncertainty Improve Model Performance?}

\author{%
  Christoph Koller \\
  Remote Sensing Technology Institute\\
  German Aerospace Center\\
  Weßling, Germany\\
  Data Science in Earth Observation\\
  Technical University of Munich\\
  Munich, Germany\\
  \texttt{christoph.koller@dlr.de} \\
  \And
  Göran Kauermann\\
  Department of Statistics\\
  LMU Munich\\
  Munich, Germany\\
  \texttt{goeran.kauermann@lmu.de}\\
  \And
  Xiao Xiang Zhu \\
  Remote Sensing Technology Institute\\
  German Aerospace Center\\
  Weßling, Germany\\
  Data Science in Earth Observation\\
  Technical University of Munich\\
  Munich, Germany\\
  \texttt{xiaoxiang.zhu@dlr.de} \\
}

\begin{document}

\maketitle

\begin{abstract}
Technological and computational advances continuously drive forward the broad field of deep learning. In recent years, the derivation of quantities describing the uncertainty in the prediction - which naturally accompanies the modeling process - has sparked general interest in the deep learning community. Often neglected in the machine learning setting is the human uncertainty that influences numerous labeling processes. As the core of this work, label uncertainty is explicitly embedded into the training process via distributional labels. We demonstrate the effectiveness of our approach on image classification with a remote sensing data set that contains multiple label votes by domain experts for each image: The incorporation of label uncertainty helps the model to generalize better to unseen data and increases model performance. Similar to existing calibration methods, the distributional labels lead to better-calibrated probabilities, which in turn yield more certain and trustworthy predictions. 
\end{abstract}

\section{Introduction}

Over the past years, deep learning has had a tremendous impact on many research fields across almost all domains. Optimized machine learning algorithms, as well as deep neural networks in particular, have enhanced the precision and accuracy of task-solving models by large margins. Concurrent advancing computational power of modern hardware has enabled such complex models to be trained while using continually shrinking time and resources. While the overall performance can be pushed by ever more complex models, the reliability of the resulting predictions is often neglected. Especially for image classification tasks, where prediction takes the form of a probability distribution over a set of possible classes, it is of crucial importance to rely on the model's confidence in its predictions. A common aspect regarding the labels is label noise \citep{frenay2013classification}, which describes the pollution of the labels from potentially various sources. Although it is argued that deep learning models are not as prone to label noise \citep{rolnick2017deep}, and calibration techniques can furthermore help to reduce the negative impact \citep{lukasik2020does}, the underlying problem remains \citep{algan2021image}. Another aspect of label quality is ambiguity among the classes, which can occur naturally in the cases of, for example, multi-label classification or head pose estimation. As a proposed solution to deal with label ambiguity, Label Distribution Learning \citep{geng2013facial, geng2016label,gao2017deep} was introduced. The framework combines the idea of labels taking the form of a distribution across the space of the different possible classes, and a suitable loss function to learn this distribution. 

Opening up the learning task to distributional labels can be highly beneficial for many applications. For safety-critical fields such as medical image analysis \citep{zhou2021review} in particular, predictions are required to be well-calibrated, while there is occasionally inevitable human label uncertainty \citep{dgani2018training}. The same problem occurs for remote sensing data, which is our application example in this paper. Label ambiguity is a known but rarely tackled problem, where examples of exception are given by using the ambiguity information for Synthetic Aperture Radar (SAR) image segmentation \citep{wang2017unsupervised} or the application of Label Distribution Learning toward aerial scene classification \citep{luo2021neighbor}. In this work, we focus on the classification of Local Climate Zones (LCZs) from satellite images. LCZs are here adopted to allocate urban conglomerates around the world into 17 different clusters \citep{stewart2012local}, which is helpful in, e.g. identifying possible urban heat islands. 

In a recently published work \citep{zhu2020so2sat} the So2Sat LCZ42 data set was introduced as a new benchmark data set for which a label confidence of 85\% was stated. We study this data set, which contains satellite images of European cities as well as additional urban areas around the globe. As a peculiarity, for each image we are presented with 10 label votes from human remote sensing experts. These votes do not coincide for many images, which reflect the inherent human uncertainty about the labels in the data and which show the difficulties entangled with the labeling process. Our novel proposal is to explicitly embed this human uncertainty during the training of a neural network classifier and investigate its performance and confidence compared to using traditional one-hot encoded labels. Our experiments send a clear message: Explicitly incorporating the inherent human uncertainty into the training process of the model is highly beneficial for both model performance and the calibration of predictive probabilities.  

\section{Related Work}
\label{sec:related}

The goal of uncertainty quantification (UQ) in the field of machine and deep learning lies in building a model that not only provides a prediction but also a measure of certainty or confidence \citep{gawlikowski2021survey}. Generally, we can distinguish between epistemic uncertainty, which is caused by the model, and aleatoric uncertainty inherent in the data \citep{hullermeier2021aleatoric}. Epistemic uncertainty can be reduced by finding a more suitable model or architecture, whereas aleatoric uncertainty is irreducible. Methodological directions to estimate uncertainty quantities include the aggregation of multiple neural networks \citep{lakshminarayanan2017simple}, deterministic networks with distributional assumptions placed on the label space \citep{malinin2018predictive, sensoy2018evidential}, sophisticated use of dropout networks \citep{gal2016dropout}, or the Bayesian neural networks \citep{blundell2015weight, maddox2019simple}. Domain-specific applications in the remote sensing area are still rare \citep{gawlikowski2022advanced, russwurm2020model}. 

Calibration is a closely related concept, which in the context of classification tasks aims at providing more reliable predictions. In particular, the probabilities derived from a machine learner should adequately describe the certainty inherent in the prediction. Originally proposed for Support Vector Machines \citep{platt1999probabilistic}, Platt Scaling describes a parametric method that trains an additional layer to transform the predictions of a classifier into calibrated probabilities. The technique has since then been widely used in various domains \citep{niculescu2005predicting, lampert2013attribute, peng2006length}. A simplified version termed Temperature Scaling (TS) has found its way into more recent deep learning-based works \citep{hinton2015distilling, guo2017calibration, mozafari2018attended}. Using only a single parameter, the approach simply scales the softmax probabilities derived from a deep learner in order to limit overconfidence. Another famous technique is termed Label smoothing (LS) \citep{szegedy2016rethinking} and helps to overcome overconfident predictions by artificially changing the labels during training. The one-hot encoded labels are smoothed by being mixed with a uniform distribution over all labels, where the degree of smoothing is steered by a hyperparameter. Several recent works \citep{muller2019does, lukasik2020does, yuan2020revisiting} have investigated this method since.  

As a novelty for the deep learning community, human uncertainty, derived from the labeling stage of the data generation process, has been studied \citep{peterson2019human}. Specifically, a set of images was labeled by multiple people and therefore received votes for potentially different classes. This rich information was then added to the training process in order to capture human categorization peculiarities \citep{battleday2020capturing} or make the underlying classification task more robust \citep{peterson2019human}. Comparisons to more classic calibration techniques such as label smoothing were drawn in \citet{zhang2021delving}. Though closely linked, the incorporation of this additional information into the labels stands in contrast to the widely studied research direction of investigating label noise \citep{brodley1999identifying, frenay2013classification, rolnick2017deep}, which focuses more on identifying mislabeled or anomalous data, or pointing out insufficiently labeled data. 

Identifying climate zones is a common task when gathering information regarding land cover from satellite images. A popular and widely-used scheme is termed Local Climate Zones (LCZs) \citep{stewart2012local}. The different classes are defined to describe urban areas and their surroundings. Originally published for the evaluation of urban heat islands (UHIs) \citep{thomas2014analysis, alexander2014local}, the scheme has since then been widely used for various climatological applications \citep{stewart2014evaluation, quanz2018micro} and urban plannings \citep{perera2018local, lelovics2014design, yang2019local}. The community-driven project termed World Urban Database and Portal (WUDAPT) \citep{mills2015introduction} targets a global high-quality coverage with the LCZ scheme. So far, significant effort has been shown to reach this goal \citep{danylo2016contributing, bechtel2015mapping, bechtel2019generating, brousse2016wudapt}. With the advancements of Deep Learning (DL) in the remote sensing community, more complex models have been established in order to classify processed satellite imagery into LCZs. Data from both the Landsat \citep{yoo2019comparison, verdonck2017influence} and the Sentinel satellite missions \citep{hu2018feature, qiu2019local} have been successfully employed for this task. More recently, a new benchmark data set \citep{zhu2020so2sat} was introduced to the community. For a scope of 42 global cities, high-quality satellite image patches are freely provided with manually crafted LCZ labels. Recent works based on this benchmark data set explore various network architectures \citep{qiu2020multilevel} or fuse multiple data sources \citep{qiu2020fusing, gawlikowski2020fusion, feng2020dynamic}. Deep learning models trained on this benchmark have also been applied to achieve global urban LCZ maps for a better understanding of the global urban morphology \citep{zhu2022}. 

\section{Methodology}

\subsection{Data}

As a basis of this work, the So2Sat LCZ42 data set \citep{zhu2020so2sat} is analyzed, comprising approximately 400k labeled image patches of size 32 $\times$ 32 pixels linked each to an area of 320m $\times$ 320m. The labels follow the classification scheme introduced by \citet{stewart2011local}: 17 characteristic climate zones, of which 10 are defined as urban zones and 7 are vegetation zones. In the publicly available version,\footnote{\href{https://mediatum.ub.tum.de/1613658}{So2Sat LCZ42 Data Set (Link to Download: \texttt{https://mediatum.ub.tum.de/1613658})}} three splits for training, validation, and testing can be chosen: A completely random split of the data, a split at the city level where each city is separated into geographically separated training and testing data, and a third approach that sets aside 10 cities from different cultural zones for testing. The labeling process was performed in a manual and labor-intensive process, which is explained in more detail by \citet{zhu2020so2sat} and largely follows a classic procedure initially used in the World Urban Database (WUDAPT) project \citep{ching2018wudapt}. Overall, 13 spectral bands of the Sentinel satellite mission with varying spatial resolutions are available from the Copernicus Hub. For the LCZ data, all 4 bands with a ground sampling distance (GSD) of 10m were chosen, as well as the bands with a GSD of 20m, which were upsampled to 10m GSD. 

\begin{figure}[h]
    \vskip 0.1in
    \centering
    \includegraphics[width=0.8\textwidth]{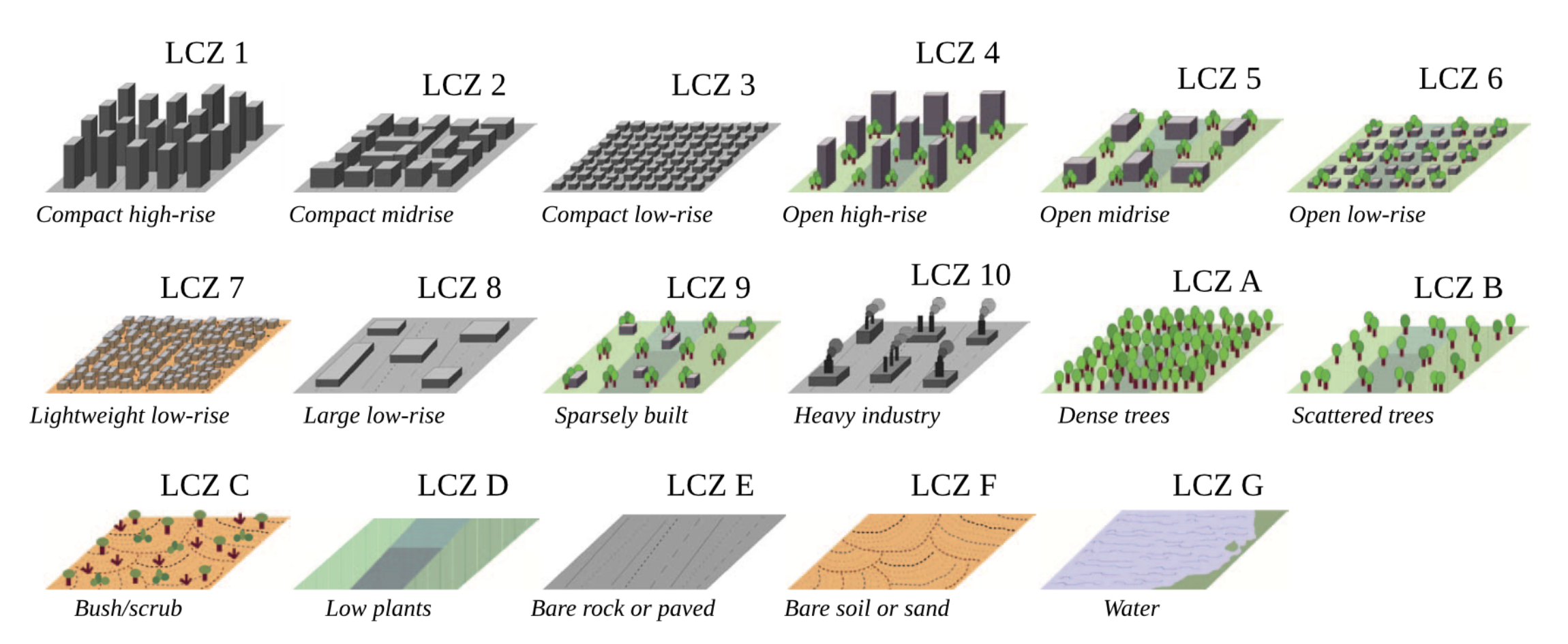}
    \caption{LCZ classification scheme for the So2Sat LCZ42 data set as shown by \citet{samsonov2017towards}. Classes 1 to 10 are urban areas, classes A to G are vegetation zones. }
    \label{fig:lcz}
\end{figure}

\subsection{From Human Votes to Labels}

As an additional experiment carried out in their work, \citet{zhu2020so2sat} launched an evaluation phase to assess the quality of the labeling process. For this, a subset of 10 European cities was chosen to serve as reference data for the labelers; 9 additional non-European regions were included to ensure a minimum level of class balance. Overall, the evaluation data set contains roughly 250k satellite image patches, which were cropped out of polygons from homogeneous regions. A group of 10 human remote sensing experts independently cast a label vote for each of the polygons, resulting in a final data set with 10 expert votes for each image. Transforming these expert votes into suitable labels for a classification task can be handled in a variety of ways. Thus, let $\bm{Y} = \bm{Y}^{(1)},\dots,\bm{Y}^{(n)}$ initially be the vote count vectors for the images $i=1,\dots,n$, where $ \bm{Y}^{(i)} = (Y_1^{(i)},...,Y_K^{(i)})$ stores the vote counts for each of the $K$ local climate zone classes for image $i$ where $K=17$. By indexing the different remote sensing experts via $j=1,\dots,J$, for image $i$ we receive the expert votes \begin{equation}
    V_1^{(i)},\dots,V_J^{(i)} \;, \; V_j^{(i)}\in \{1,\dots,K\}\,\, \forall \, i = 1,\dots,n \,.
\end{equation} We define the votes as a $K$-dimensional vector $\bm{V}_j^{(i)} = (\mathbb{1}_{\{V_j^{(i)}=1\}},\dots,\mathbb{1}_{\{V_j^{(i)}=K\}})$ and obtain the vote counts via $Y_k^{(i)}=\sum_j \mathbb{1}_{\{V_j^{(i)}=k\}}$. In particular, $Y_k^{(i)} = m$ means that for image $i$  class $k$ received $m$ votes and it holds that $ \sum_{k=1}^KY_k^{(i)} = M$, where $M=10$ represents the number of votes or experts. A common strategy is to rely on the majority vote of the experts that we define for image $i$ as $Y_{max}^{(i)} := \max\limits_j Y_j^{(i)}$ to be the class which received the most expert votes. The associated one-hot encoded label is denoted as a $K$-dimensional vector \begin{equation}
    \begin{split}
            \bm{y}^{(i)}_{\text{max}} = (\mathbb{1}_{\{Y_1^{(i)}=Y_{max}^{(i)}\}}, \dots,\mathbb{1}_{\{Y_K^{(i)}=Y_{max}^{(i)}\}})\;, \text{where}\;\mathbb{1}_{\{Y_j^{(i)}=Y_{max}^{(i)}\}} = 1 \Leftrightarrow Y_j^{(i)}=Y_{max}^{(i)} \, .
    \end{split}
    \label{eq:majority}
\end{equation}
This simplification gives rise to the question of how much information is lost when relying on the majority label decision. It is argued by \citet{zhu2020so2sat}, that majority voting does help to improve label confidence. Yet part of the uncertainty in the voting process is hidden from the classifier when it is presented with the majority vote $\bm{y}_{\text{max}}$ in (\ref{eq:majority}). An alternative approach is hence to incorporate the entirety of all votes directly into the label by forming a distributional label. To do so, we directly use the empirical distribution formed by the observed votes and define the soft label for image $i$ in the following discussions via \begin{equation}
    \bm{y}^{(i)}_{\text{distr}}=\bm{Y}^{(i)} / \, M \label{eq:distr}
\end{equation} By doing so, the mode of the distributional label coincides with the mode of $\bm{y}_{\text{max}}$ (\ref{eq:majority}), but the distributional form allows for a more flexible learning approach, as other classes that were voted for in the evaluation process are also considered. 

\subsection{Learning Distributions}

In the following, let $\{\bm{x}^{(i)},\bm{y}^{(i)}\}_{i=1,\dots,n}\in (\mathcal{X} \times \mathcal{Y})^n$ be the classification data: $\bm{x}^{(1)}, \dots, \bm{x}^{(n)} \in \mathcal{X}$ are the multi-spectral LCZ42 image patches, and $\bm{y}^{(1)}, \dots, \bm{y}^{(n)} \in \mathcal{Y}$  are the corresponding labels. Furthermore, let $f_{\theta}(x)$ be a neural network classifier based on the parameters stored in $\theta$. Given an input $x \in \mathcal{X}$, the predictive distribution of the network is denoted by $p_{\theta}(y|x)$, and $p_{\theta}(y=k|x)$ returns the estimated probability of $x$ belonging to class $k$. For training, typically the cross-entropy loss is used in a classification setting. From an information theory perspective, the cross-entropy defines the amount of additional information needed to approximate a sample from the source distribution. However, in the literature for the recently established research field of Label Distribution Learning, the Kullback-Leibler (KL) divergence has been established as a loss function \citep{geng2016label, geng2013facial, gao2017deep}. Given a true distribution, the KL divergence measures the information lost when the target distribution is used as an approximation. It measures the dissimilarity between two probability functions and is therefore also termed relative entropy. 

If we now assume for the So2Sat LCZ42 data that the distribution formed by the votes of the 10 remote sensing experts is the ground truth label distribution, we can measure the dissimilarity between the predicted and the ground truth distribution with the KL divergence. When implemented as a loss function, it is derived for a batch of data $\{\bm{x}^{(i)},\bm{y_{\text{distr}}}^{(i)}\}_{i=1,\dots,,m}$ via   

\begin{align*}
    \mathcal{L}_{KL}(f_{\theta},\bm{x}^{(1)},\dots,\bm{x}^{(m)},\bm{y_{\text{distr}}}^{(1)},\dots,\bm{y_{\text{distr}}}^{(m)}) = -\frac{1}{m} \displaystyle\sum_{i=1}^m\displaystyle\sum_{k=1}^K y_{\text{distr}, k}^{(i)} \cdot \log \frac{y_{\text{distr}, k}^{(i)}}{p_{\theta}(\bm{y}^{(i)}=k|\bm{x}^{(i)})} \, . 
\end{align*} Note that for $y_{\text{distr}, k}^{(i)} = 0 $ we set $y_{\text{distr}, k}^{(i)} \log(y_{\text{distr}, k}^{(i)}) \equiv 0$. This is done to measure the information loss when using the predictive distribution of the neural network to approximate the assumed ground truth distribution over labels. Consequently, training with the KL divergence describes the process of iteratively finding a neural network mimicking human voting behavior. 

\subsection{Assessing the Quality of Predictive Uncertainty}

Neural network classifiers are often prone to overconfidence, namely that predicted probabilities for a class overestimate the percentage of times the algorithm actually yields a correct prediction \citep{guo2017calibration}. One then speaks
of poorly calibrated predictions, as they do not represent a well-defined predictive probability estimate. In this circumstance, derived uncertainty quantities are not reliable, because the underlying probabilities are ill-defined in the first place. Therefore, one can speak of a frequentist notion of uncertainty when referring to calibration \citep{lakshminarayanan2017simple}. Deviations from the perfectly calibrated model can be measured with different error rates, of which the Expected Calibration Error (ECE) is the most prominent one. This calibration error gives a good implication of whether the probabilities predicted by a network are a good indicator of the true uncertainty affiliated with the prediction. It can be further visualized in a so-called reliability diagram \citep{degroot1983comparison, niculescu2005predicting}, which displays the accuracy versus the corresponding confidence in a single plot, averaged over predefined intervals. Following the notation of \citet{guo2017calibration}, we denote these bins by $B_m = \big(\frac{m-1}{M},\frac{m}{M}\big]\;, \; m=1,\dots,M$ and define the indices of the images whose confidences $\hat{p}^{(i)}$ fall into the respective bin by $I_m\;, \; m=1,\dots,M$. The needed quantities are then calculated by \begin{equation*}
    \text{acc}(I_m) = \displaystyle\sum_{i \in I_M} \mathbb{1}_{\{ \hat{y}^{(i)} =  \bm{y}_{\text{max}}^{(i)}\}} \;\; \text{and} \;\; \text{conf}(I_m)=\displaystyle\sum_{i \in I_m}\hat{p}^{(i)} 
\end{equation*}
and displayed in a 2-dimensional plot with respect to the earlier defined intervals. Here, $\bm{\hat{y}}^{(i)}$ is the one-hot encoded predicted class for input $\bm{x}^{(i)}$, $\hat{p}^{(i)}$ the corresponding predicted probability, and one defines $\bm{y}_{\text{max}}^{(i)}$ as in (\ref{eq:majority}). For a 10-class classification problem, one could therefore set $M=10$, for example, which would lead to $B_1 = (0,0.1], B_2 = (0.1,0.2]$ and so on. Exemplary, the interval $B_2$ includes all images with confidences ranging between 0.1 (excluding) and 0.2 (including). The ECE is then derived via \begin{equation*}
    \text{ECE} = \displaystyle\sum_{m=1}^M \frac{|I_m|}{n} |\text{acc}(I_m) - \text{conf}(I_m)| \, .
\end{equation*}

\section{Experiments}

The introduced quantities allow for a direct comparison between two main ways of modeling the So2Sat LCZ42 data set introduced above, given the independent expert votes. In a more general setting, the final one-hot encoded label is derived by taking the majority vote of the experts. Here, training is performed in the usual manner using the cross-entropy loss. In a more adaptive way, we can make explicit use of the distribution over labels formed by the expert votes in order to account for the inherent uncertainty within the labeling process. By doing so, we implicitly assume the underlying label distribution of the votes to serve as a ground truth label, or at least to better reflect the ground truth distribution. Similar to \citet{peterson2019human}, we state that by incorporating human uncertainty in form of the distributional label, it is possible to find a better estimator for the predictive distribution $p_{\theta}(y|x)$ of the network classifier. The network training then differs in the specification of the label and the loss; the KL divergence is used to better reflect the set goal of approximating the true distributional label formed by the expert votes. We furthermore draw the comparison to state-of-the-art calibration methods. 

\subsection{Setup and Settings}

For the classification task, we use the model introduced by \citet{qiu2020multilevel} (Sen2LCZ-Net), which is a modified Convolutional Neural Network (CNN) using intermediate deep feature representations at multiple stages of the network. These representations are then averaged and pooled at the end before being transformed into the logit space. The model has been shown to be superior over many state-of-the-art neural network architectures in extensive benchmark tests on the So2Sat LCZ42 data set by the authors. The overall disagreement among voters in the non-urban classes A-G is lower than for the urban classes (see Appendix A.2 for more information). In the following, we therefore focus our analysis on the urban classes 1 through 10 in order to place a stronger focus on learning with distributional labels. The question we aim to answer is whether the use of label distributions helps to significantly boost model calibration and performance. An important aspect of this question aims toward the ability of the model to generalize to unseen data. In particular, this data does not necessarily need to be from the exact same domain or distribution as the data seen during training. To account for this, the data set was divided on a city level into a geographically separated training and testing data set. The latter was then randomly split into data for validation and testing. See Appendix A.1 for more details. 

\subsection{Results}

Sen2LCZ-Net was trained on the training set utilizing different labels and loss functions plus additional off-the-shelf calibration methods. The categorical cross-entropy loss incorporates the one-hot encoded labels based on the majority vote of the expert votes, and the KL divergence takes the distribution formed by the empirical distribution of the expert votes. All models were trained using the training data and validation data specified in Appendix A.1. Identical hyperparameters were used (after grid search), and the same stopping and convergence criteria were set. For more implementation details see Section A.5 in the Appendix. In particular, the same input in the form of satellite imagery was used in both models. 

\begin{figure*}[h]
    \vskip 0.1in
    \subfloat[One-Hot Encoding]{
    \includegraphics[width=0.48\textwidth]{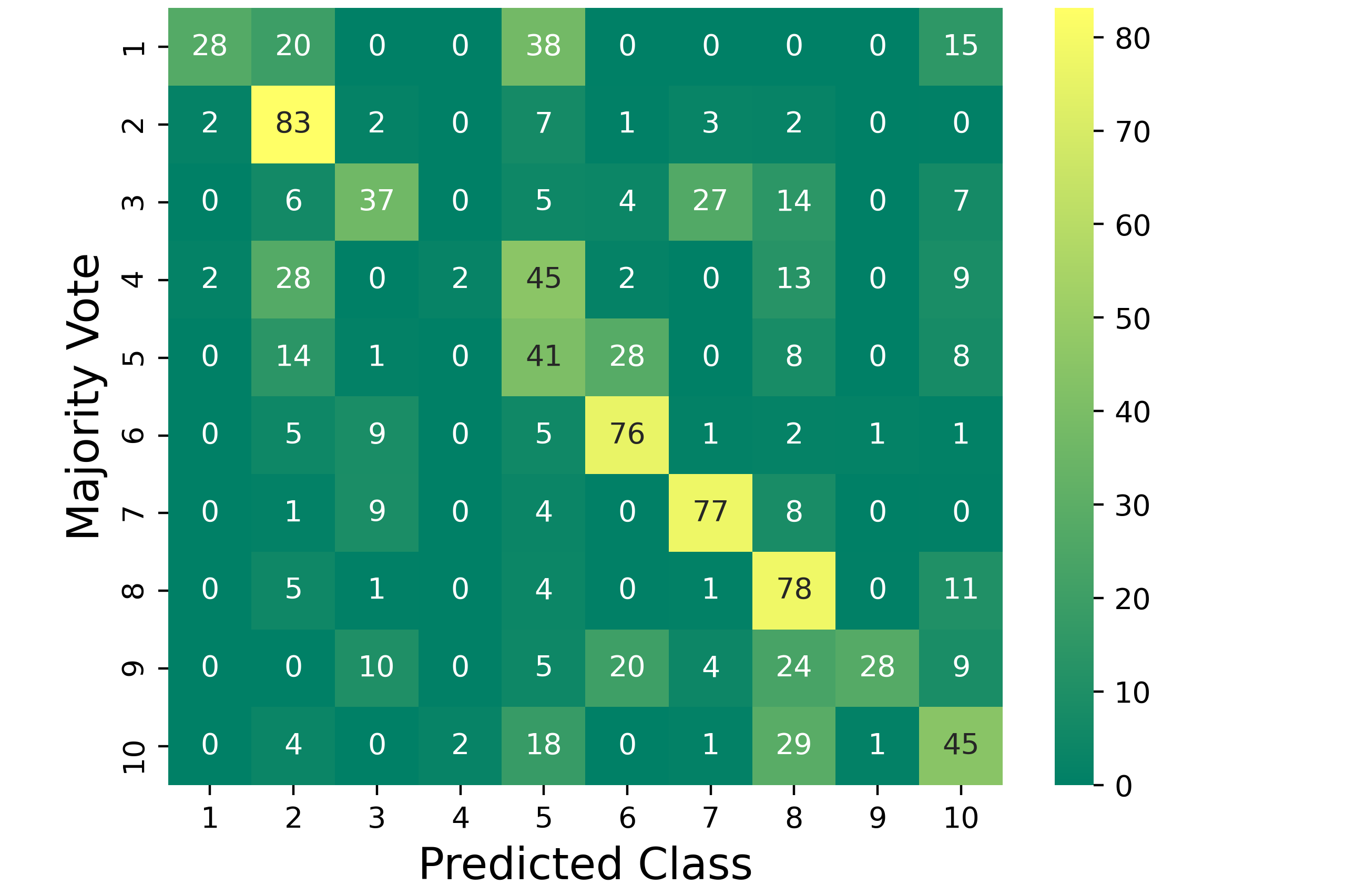}
    }
    \qquad
    \subfloat[Label Distribution Encoding]{
    \includegraphics[width=0.48\textwidth]{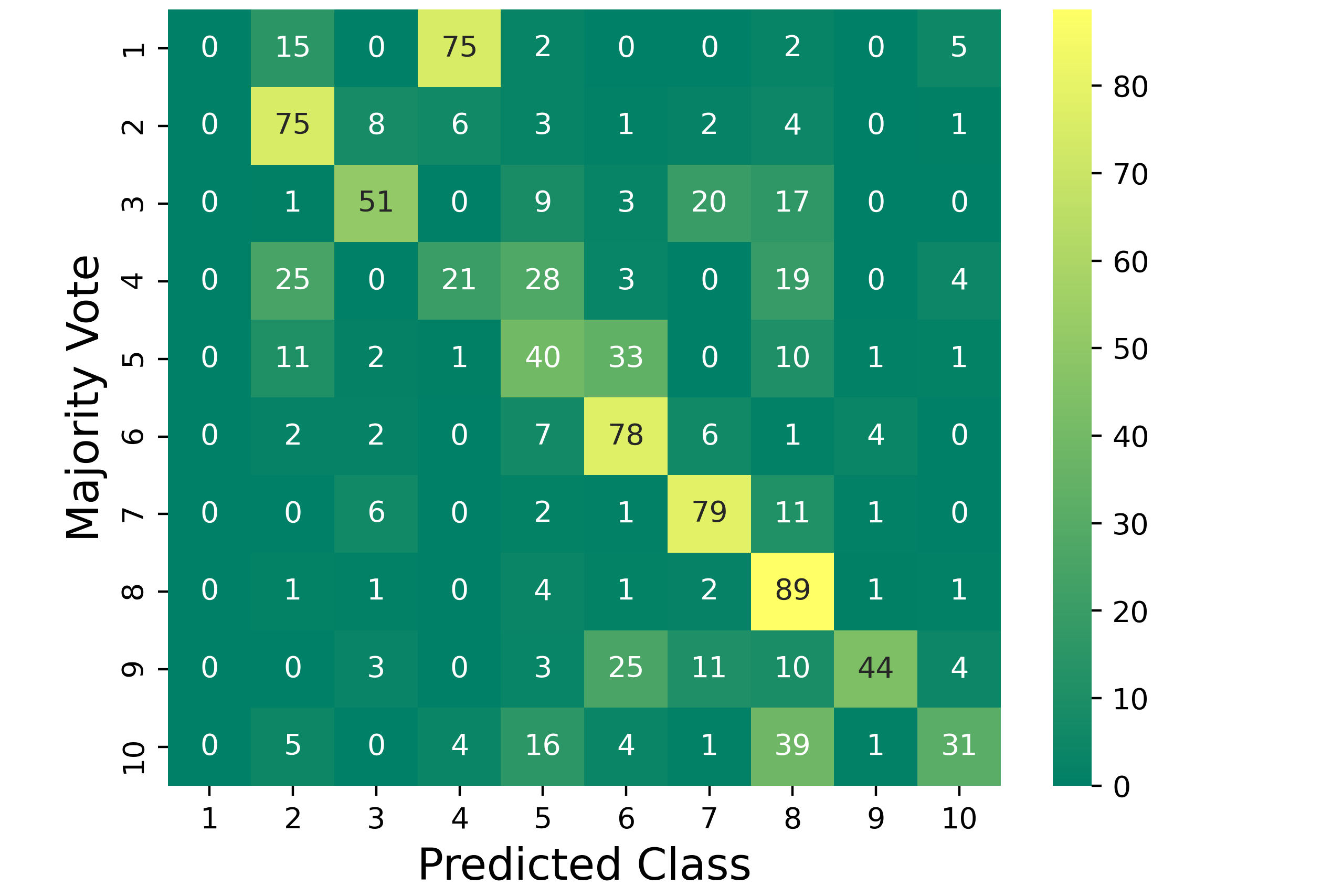}
    }
    \caption{Exemplary pairwise confusion between predicted class and respective majority vote for the test set. }
    \label{fig:confusion_test}
\end{figure*}

Exemplary confusion matrices of the predictions on the test set are presented for both models in Figure \ref{fig:confusion_test}. Here, the ground truth label is set to be the majority vote of the experts, and the predicted class is defined via the highest predicted probability of the respective network $\hat{p} = \max_k p_{\theta} \,(y = k|x)$. First note that for most of the part, both models have the same pitfalls of misclassifying certain classes. This holds in particular for class 3 (compact low-rise) being falsely classified as class 7 (lightweight low-rise), class 4 (open high-rise) being misclassified as class 2 (compact midrise), as well as class 5 (open midrise) being classified as class 6 (open low-rise). Adding to this claim is the strong confusion with class 8 (large low-rise), which is often falsely predicted for images from various classes (with respect to the majority vote). In a similar manner, images for which experts agreed on class 9 (sparsely built) are often falsely classified as classes 6 and 8. The results regarding the majority-voted class 1 seem rather arbitrary, however, as they largely depend on the data distribution of the few test samples. Of particular interest is furthermore the comparison between the model confusion and the confusion among voters. With one in every three votes deviating from the majority vote, experts chose to vote for class 3 in cases where the majority of experts settled for class 7. This large confusion is only slightly reflected in model confusion. The opposite case is matched rather closely in the predictions of both models on the test set. With more than one in every six votes deviating from the majority vote, the combinations of classes 4 and 5, as well as 9 and 6, are also found to be confusing for the model in both training cases. However, in the cases of classes 4 and 5, it seems to be significantly less confusing for the model trained with distributional labels. Further deviations of the model confusion from the human confusion can be partially attributed to the previously mentioned arbitrariness of the very low number of affected samples in the data.  All other model confusions cannot be directly linked to the confusion in the voting process and therefore result from difficulties entirely attributable to the model.

\begin{table}[h] \small
     \caption{Performance scores on the test set. Accuracy and other related measures were derived with respect to the majority vote. All scores are averaged over 5 runs. OA = overall accuracy, MAA = macro avg. accuracy, WAA = weighted avg. accuracy, $\kappa$ = kappa score, LS = Label smoothing. }
    \centering
    \begin{tabular}{lcccc}
    \toprule \addlinespace
    & OA $\uparrow$ & MAA  $\uparrow$ & WAA  $\uparrow$ & $\kappa$  $\uparrow$ \\ \addlinespace
    One-hot & 68.4 $\pm$ 5.5 & 42.9 $\pm$ 6.4 & 69.6 $\pm$ 2.2 & 60.2 $\pm$ 6.4 \\ 
    + LS & 67.5 $\pm$ 2.4 & \bm{$50.3 \pm 3.5$} & 69.9 $\pm$ 2.0 & 59.4 $\pm$ 2.7 \\ \addlinespace
    Distr. & 67.0 $\pm$ 2.2 & 45.8 $\pm$ 3.9 & \bm{$71.0 \pm 0.5$} & 58.8 $\pm$ 2.4 \\ 
    + LS & \bm{$68.6 \pm 2.3$} & 43.4 $\pm$ 6.1 & 69.7 $\pm$ 2.1 & \bm{$60.4 \pm 2.8$} \\ \addlinespace 
    \bottomrule
    \end{tabular}
    \label{table:performance}
\end{table}

We can deduce the results of the trained models in Table \ref{table:performance}. For the distributional label approach, accuracy was again measured with respect to majority voting, which helps to explain the observed minor differences in performance between the two model configurations. Although the distributional approach performs on average better than the regular approach in 3 out of 4 metrics, the macro average accuracy lacks behind by a large margin. Regarding the deviance of predictions from the true labels, Table \ref{table:ce20} shows the cross-entropy between the two distributions on the identical test set as well as several calibration error rates (based on 20 identically-sized bins). \begin{table}[h!]
    \small
    \caption{Cross-Entropies between predicted softmax probabilities and labels on the test set as well as calibration errors, averaged over five runs. CE = Cross entropy, LS = Label smoothing, TS = Temperature Scaling, MC-Drop = Monte Carlo Dropout. Binning was performed using 20 equally-sized bins. }
    \centering
    \begin{tabular}{lccccc}
    \toprule \addlinespace
    & CE One-hot $\downarrow$ & CE Distr. $\downarrow$ & ECE $\downarrow$ & MCE $\downarrow$ & SCE $\downarrow$ \\ \addlinespace
    \cmidrule{2-6} 
    One-hot & 1.12 $\pm$ 0.05 & 1.38 $\pm$ 0.07 & 9.79 $\pm$ 3.18 & 23.14 $\pm$ 3.97 & \bm{$1.03 \pm 0.50$} \\ 
    + LS & 1.05 $\pm$ 0.01 & 1.23 $\pm$ 0.03 & 7.33 $\pm$ 2.62 & 19.80 $\pm$ 4.82 & 1.11 $\pm$ 0.25 \\ 
    + TS & 1.00 $\pm$ 0.13 & 1.17 $\pm$ 0.07 & 4.15 $\pm$ 2.37 & 15.88 $\pm$ 10.60 & 1.44 $\pm$ 0.22 \\ 
    + LS \& TS & 1.02 $\pm$ 0.03 & 1.18 $\pm$ 0.02 & \bm{$3.21 \pm 0.96$} & \bm{$11.30 \pm 4.26$} & 1.32 $\pm$ 0.03 \\
    + MC-Drop & 1.12 $\pm$ 0.05 & 1.37 $\pm$ 0.06 & 9.57 $\pm$ 3.11 & 23.10 $\pm$ 4.22 & 1.04 $\pm$ 0.49 \\
    + LS \& MC-Drop & 1.05 $\pm$ 0.01 & 1.23 $\pm$ 0.03 & 7.11 $\pm$ 2.41 & 33.22 $\pm$ 26.60 & 1.12 $\pm$ 0.24 \\ \addlinespace 
    Distr. & $1.06\pm0.07$ & $1.21\pm0.07$ & 5.80$\pm$1.07 & $15.57 \pm 4.22$ & 1.21 $\pm$ 0.20 \\ 
    + LS & $0.98\pm0.03$ & $1.08\pm0.02$ & $8.31\pm 2.46$ & $17.32\pm4.84$ & 1.73 $\pm$ 0.06 \\ 
    + TS & 0.96 $\pm$ 0.09 & 1.07 $\pm$ 0.07 & 5.89 $\pm$ 2.50 & 15.37 $\pm$ 3.94 & 1.72 $\pm$ 0.15 \\ 
    + LS \& TS & \bm{$0.95 \pm 0.04$} & \bm{$1.05 \pm 0.04$} & 4.21 $\pm$ 1.38 & 15.35 $\pm$ 1.95 & 1.58 $\pm$ 0.10 \\ \addlinespace 
    \bottomrule
    \end{tabular}
    \label{table:ce20}
\end{table} Next to the earlier derived ECE, we show the Maximum Calibration Error (MCE)\citep{guo2017calibration} and the Static Calibration Error (SCE)\citep{nixon2019measuring}. The former error only considers the maximum of all bin-wise discrepancies between accuracy and confidence, whereas the latter weighs all derived metrics with respect to the class frequencies. For methodological and implementation details see Appendices A.3 and A.5, respectively. When trained with distributional labels, the model can fit better toward the test label distributions when predicting on the test set by a large margin. This holds true even though the model with distributional labels was trained with the KL divergence as loss. While label smoothing helps to improve the performance in both settings, temperature scaling leads to mixed calibration performance results. The additional use of Monte Carlo Dropout was enforced via averaging the confidences of multiple predictions with activated dropout. While this enabled to bound the ECE to a reasonable level, the generalization performance was nearly unaffected. Overall, all calibration errors benefit from the uncertainty-guided approach, although the overall best errors are achieved with conventional calibration techniques.

Yet more unforeseen is the cross-entropy with respect to the one-hot encoded labels. Although the model trained with the distributional labels uses a different loss function and different labels, it can better approximate the one-hot encoded test labels and underlines the strong generalization performance of the human uncertainty models. While the additional off-the-shelf calibration methods help to marginally improve the cross-entropies, the ECE is negatively affected by individual calibration methods in the distributional setting. An explanation for this could be the fact that the optimization of the temperature is prone to overfitting on the validation data set, leading to a poor generalization when already considering human uncertainty within the training process. The overall better cross-entropies of the models including human label uncertainty underline a crucial aspect of training with distributional labels: If chosen appropriately, they keep the model more flexible when generalizing to unseen data. When explicitly modeling label distributions, all classes with a nonzero entry in the distribution of the label are affected by the loss function; therefore the training is more flexible and less prone to overconfidence in its predictions. 

\begin{figure*}[h]
    \vskip 0.1in
    \subfloat[One-Hot Encoding]{
    \includegraphics[width=0.46\textwidth]{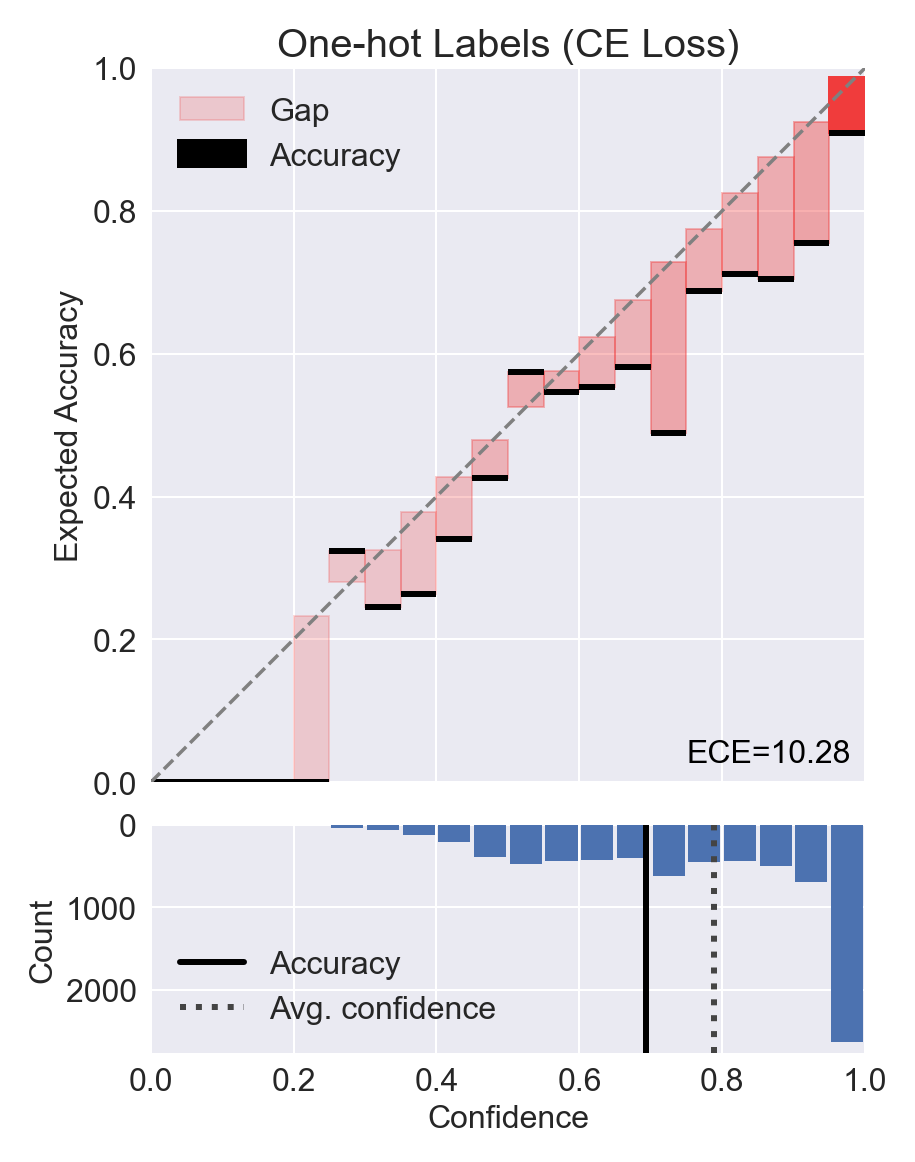}
    }
    \qquad
    \subfloat[Label Distribution Encoding]{
    \includegraphics[width=0.46\textwidth]{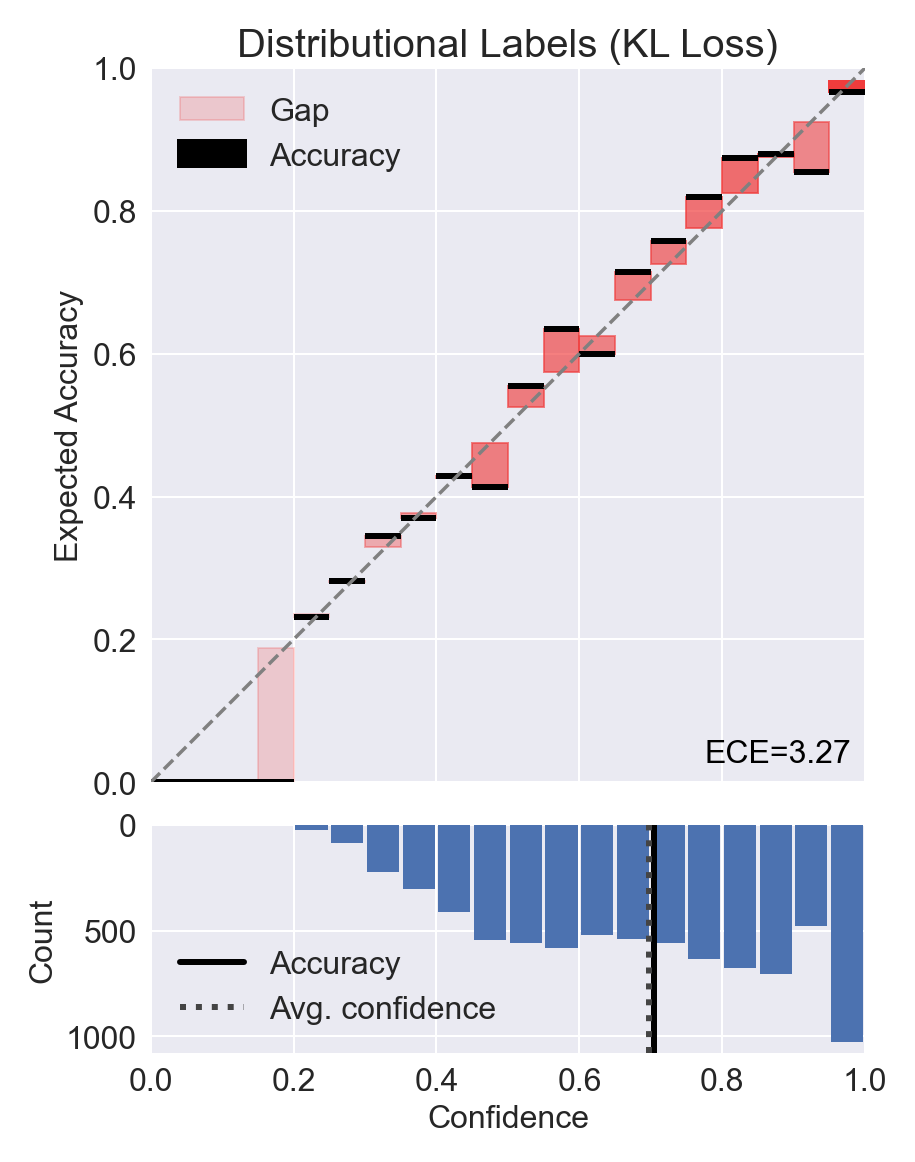}
    }
    \caption{Exemplary reliability diagrams on the test set. Both visualizations were created using \citet{reliabilitydiagram}.}
    \label{fig:reliability_test}
\end{figure*}

This phenomenon is particularly visible in Figure \ref{fig:reliability_test}, which displays reliability diagrams \citep{guo2017calibration} of a single run for the two modeling approaches in an uncalibrated setting. Clearly visible is the significantly lower ECE for the model trained with the distributional labels, especially for the regions with higher confidence. In these regions, the confidence is more meaningful and is reflected better by the accuracy, resulting in a better-calibrated model when using distributional labels. Since calibration can be treated as the frequentist notion of uncertainty \citep{lakshminarayanan2017simple}, we can deduce that by incorporating the voting uncertainty into the training process, the resulting model is better calibrated and hence yields a predictive distribution that contains a more feasible sense of uncertainty. 

Moreover, the lower part of Figure \ref{fig:reliability_test} shows the overconfidence of the one-hot encoded model, where the average confidence in the predictions exceeds the overall accuracy by a large margin. On the contrary, the accuracy is almost precisely met by the average confidence when the model has been trained with distributional labels. Note, however, that here the accuracy was again calculated by means of the majority vote when modeling the distributional labels. This can lead to underconfidence (as opposed to overconfidence in the one-hot case) because a set of images can be classified correctly even with a relatively low average confidence if the label distributions are multimodal. 

\section{Discussion}
\label{sec:discussion}

Generally speaking, it is desirable to find a suitable model calibration technique in order to avoid under- or overconfidence of the model in its predictions. Traditional methods such as temperature scaling \citep{platt1999probabilistic, niculescu2005predicting} or label smoothing \citep{szegedy2016rethinking} train additional hyperparameters using the validation data set for this purpose. The approach we take here is vastly different, though it achieves similar goals, and is closely related to the works of \citet{peterson2019human}. We claim that incorporating human uncertainty about the labels helps to overcome not only the poor calibration of the predicted probabilities but also the inability to generalize well on data not seen during training. Given the special structure of the studied data set, the label votes cast by 10 remote sensing experts are used to form an empirical distribution over the classes that serves as an approximation of the inherent human label uncertainty. This distribution is then embedded as a distributional label for training. Due to its theoretical properties, KL divergence is implemented as a loss function. In order to explicitly measure the impact of human uncertainty, the model architecture, as well as affiliated hyperparameters, remained unchanged. 

The questions we pose while forming the distributional labels are whether this approximation (a) is sufficient and (b) has a potential benefit for modeling purposes and downstream tasks. The latter can be partially answered by the findings of this work: we see a direct improvement in terms of calibration regarding the predictive distribution when incorporating the distributional labels for the uncalibrated baseline model. Here, the ECE was on average almost halved. Also with regard to the overall quality of the approximation, the model benefited from the label distributions. An improvement of approximately 10\% on average in the cross-entropy could be seen at test time. This holds for the cross-entropy both between the predicted probabilities and the ground truth label distributions and between the probabilities and the one-hot encoded labels. We see huge potential, particularly in the application to safety-critical domains that rely heavily on well-calibrated probabilities. As for the first question posed, one has to bear in mind that labeling images is a labor-intensive process. This holds especially true when there is a need for experts in the field, as in the studied case for the LCZ42 data set \citep{zhu2020so2sat}, but also in fields such as medical image analysis or real-time object detection. A natural question is therefore whether it would be more beneficial to have a larger number of images or individual votes overall. We see this question as a very promising research direction and will leave it for future work.  

\section{Conclusion}

In this work, we examine the impact of employing human uncertainty in the training process of a neural network classifier. We apply our methodology to the So2Sat LCZ42 data, where the images stem from the Sentinel satellite mission and cover several large European cities as well as additional areas from all over the world. Ten expert label votes are supplied for each satellite image, which comprise a notion of uncertainty within the classification task. Forming a label distribution from these votes allows us to directly implement the human uncertainty linked to the voting process into the network training. To do so, we employ the framework of Label Distribution Learning, which enables the model to better adapt to the uncertainty rooted in the labels. 

When label uncertainty is incorporated into training, an improvement in the generalization performance of the model can be measured for the remote sensing data set studied. The overall loss when generalizing to unseen data was reduced on average by a margin of approximately 10\%. Off-the-shelf calibration methods such as label smoothing or temperature scaling lead to improved performance competitive to the uncertainty-guided approach, yet while requiring additional data for hyperparameter tuning. The ECE, a key measure of calibration quality, is almost halved in the majority of experiments by means of the embedded human label uncertainty when no further calibration technique is applied. The improved calibration and generalization performance of the uncertainty-guided approach can be directly converted to a more feasible notion of predictive uncertainty. This is because the predictive probabilities of the network, when generalizing to unseen data, yield more reliable estimates of the unknown label distributions associated with the data.  

\newpage

\small

\bibliography{references}

\begin{thebibliography}{65}
\providecommand{\natexlab}[1]{#1}
\providecommand{\url}[1]{\texttt{#1}}
\expandafter\ifx\csname urlstyle\endcsname\relax
  \providecommand{\doi}[1]{doi: #1}\else
  \providecommand{\doi}{doi: \begingroup \urlstyle{rm}\Url}\fi

\bibitem[Alexander \& Mills(2014)Alexander and Mills]{alexander2014local}
Alexander, P.~J. and Mills, G.
\newblock Local climate classification and dublin’s urban heat island.
\newblock \emph{Atmosphere}, 5\penalty0 (4):\penalty0 755--774, 2014.

\bibitem[Algan \& Ulusoy(2021)Algan and Ulusoy]{algan2021image}
Algan, G. and Ulusoy, I.
\newblock Image classification with deep learning in the presence of noisy
  labels: A survey.
\newblock \emph{Knowledge-Based Systems}, 215:\penalty0 106771, 2021.

\bibitem[Battleday et~al.(2020)Battleday, Peterson, and
  Griffiths]{battleday2020capturing}
Battleday, R.~M., Peterson, J.~C., and Griffiths, T.~L.
\newblock Capturing human categorization of natural images by combining deep
  networks and cognitive models.
\newblock \emph{Nature communications}, 11\penalty0 (1):\penalty0 1--14, 2020.

\bibitem[Bechtel et~al.(2015)Bechtel, Alexander, B{\"o}hner, Ching, Conrad,
  Feddema, Mills, See, and Stewart]{bechtel2015mapping}
Bechtel, B., Alexander, P.~J., B{\"o}hner, J., Ching, J., Conrad, O., Feddema,
  J., Mills, G., See, L., and Stewart, I.
\newblock Mapping local climate zones for a worldwide database of the form and
  function of cities.
\newblock \emph{ISPRS International Journal of Geo-Information}, 4\penalty0
  (1):\penalty0 199--219, 2015.

\bibitem[Bechtel et~al.(2019)Bechtel, Alexander, Beck, B{\"o}hner, Brousse,
  Ching, Demuzere, Fonte, G{\'a}l, Hidalgo, et~al.]{bechtel2019generating}
Bechtel, B., Alexander, P.~J., Beck, C., B{\"o}hner, J., Brousse, O., Ching,
  J., Demuzere, M., Fonte, C., G{\'a}l, T., Hidalgo, J., et~al.
\newblock Generating wudapt level 0 data--current status of production and
  evaluation.
\newblock \emph{Urban climate}, 27:\penalty0 24--45, 2019.

\bibitem[Blundell et~al.(2015)Blundell, Cornebise, Kavukcuoglu, and
  Wierstra]{blundell2015weight}
Blundell, C., Cornebise, J., Kavukcuoglu, K., and Wierstra, D.
\newblock Weight uncertainty in neural network.
\newblock In \emph{International Conference on Machine Learning}, pp.\
  1613--1622. PMLR, 2015.

\bibitem[Brodley \& Friedl(1999)Brodley and Friedl]{brodley1999identifying}
Brodley, C.~E. and Friedl, M.~A.
\newblock Identifying mislabeled training data.
\newblock \emph{Journal of artificial intelligence research}, 11:\penalty0
  131--167, 1999.

\bibitem[Brousse et~al.(2016)Brousse, Martilli, Foley, Mills, and
  Bechtel]{brousse2016wudapt}
Brousse, O., Martilli, A., Foley, M., Mills, G., and Bechtel, B.
\newblock Wudapt, an efficient land use producing data tool for mesoscale
  models? integration of urban lcz in wrf over madrid.
\newblock \emph{Urban Climate}, 17:\penalty0 116--134, 2016.

\bibitem[Ching et~al.(2018)Ching, Mills, Bechtel, See, Feddema, Wang, Ren,
  Brousse, Martilli, Neophytou, et~al.]{ching2018wudapt}
Ching, J., Mills, G., Bechtel, B., See, L., Feddema, J., Wang, X., Ren, C.,
  Brousse, O., Martilli, A., Neophytou, M., et~al.
\newblock Wudapt: An urban weather, climate, and environmental modeling
  infrastructure for the anthropocene.
\newblock \emph{Bulletin of the American Meteorological Society}, 99\penalty0
  (9):\penalty0 1907--1924, 2018.

\bibitem[Chollet et~al.(2015)]{chollet2015keras}
Chollet, F. et~al.
\newblock Keras.
\newblock \url{https://keras.io}, 2015.

\bibitem[Danylo et~al.(2016)Danylo, See, Bechtel, Schepaschenko, and
  Fritz]{danylo2016contributing}
Danylo, O., See, L., Bechtel, B., Schepaschenko, D., and Fritz, S.
\newblock Contributing to wudapt: A local climate zone classification of two
  cities in ukraine.
\newblock \emph{IEEE Journal of Selected Topics in Applied Earth Observations
  and Remote Sensing}, 9\penalty0 (5):\penalty0 1841--1853, 2016.

\bibitem[DeGroot \& Fienberg(1983)DeGroot and Fienberg]{degroot1983comparison}
DeGroot, M.~H. and Fienberg, S.~E.
\newblock The comparison and evaluation of forecasters.
\newblock \emph{Journal of the Royal Statistical Society: Series D (The
  Statistician)}, 32\penalty0 (1-2):\penalty0 12--22, 1983.

\bibitem[Dgani et~al.(2018)Dgani, Greenspan, and Goldberger]{dgani2018training}
Dgani, Y., Greenspan, H., and Goldberger, J.
\newblock Training a neural network based on unreliable human annotation of
  medical images.
\newblock In \emph{2018 IEEE 15th International Symposium on Biomedical Imaging
  (ISBI 2018)}, pp.\  39--42. IEEE, 2018.

\bibitem[Feng et~al.(2020)Feng, Lin, He, Guan, Wang, and Shi]{feng2020dynamic}
Feng, P., Lin, Y., He, G., Guan, J., Wang, J., and Shi, H.
\newblock A dynamic end-to-end fusion filter for local climate zone
  classification using sar and multi-spectrum remote sensing data.
\newblock In \emph{IGARSS 2020-2020 IEEE International Geoscience and Remote
  Sensing Symposium}, pp.\  4231--4234. IEEE, 2020.

\bibitem[Fr{\'e}nay \& Verleysen(2013)Fr{\'e}nay and
  Verleysen]{frenay2013classification}
Fr{\'e}nay, B. and Verleysen, M.
\newblock Classification in the presence of label noise: a survey.
\newblock \emph{IEEE transactions on neural networks and learning systems},
  25\penalty0 (5):\penalty0 845--869, 2013.

\bibitem[Gal \& Ghahramani(2016)Gal and Ghahramani]{gal2016dropout}
Gal, Y. and Ghahramani, Z.
\newblock Dropout as a bayesian approximation: Representing model uncertainty
  in deep learning.
\newblock In \emph{international conference on machine learning}, pp.\
  1050--1059. PMLR, 2016.

\bibitem[Gao et~al.(2017)Gao, Xing, Xie, Wu, and Geng]{gao2017deep}
Gao, B.-B., Xing, C., Xie, C.-W., Wu, J., and Geng, X.
\newblock Deep label distribution learning with label ambiguity.
\newblock \emph{IEEE Transactions on Image Processing}, 26\penalty0
  (6):\penalty0 2825--2838, 2017.

\bibitem[Gawlikowski et~al.(2020)Gawlikowski, Schmitt, Kruspe, and
  Zhu]{gawlikowski2020fusion}
Gawlikowski, J., Schmitt, M., Kruspe, A., and Zhu, X.~X.
\newblock On the fusion strategies of sentinel-1 and sentinel-2 data for local
  climate zone classification.
\newblock In \emph{IGARSS 2020-2020 IEEE International Geoscience and Remote
  Sensing Symposium}, pp.\  2081--2084. IEEE, 2020.

\bibitem[Gawlikowski et~al.(2021)Gawlikowski, Tassi, Ali, Lee, Humt, Feng,
  Kruspe, Triebel, Jung, Roscher, et~al.]{gawlikowski2021survey}
Gawlikowski, J., Tassi, C. R.~N., Ali, M., Lee, J., Humt, M., Feng, J., Kruspe,
  A., Triebel, R., Jung, P., Roscher, R., et~al.
\newblock A survey of uncertainty in deep neural networks.
\newblock \emph{arXiv preprint arXiv:2107.03342}, 2021.

\bibitem[Gawlikowski et~al.(2022)Gawlikowski, Saha, Kruspe, and
  Zhu]{gawlikowski2022advanced}
Gawlikowski, J., Saha, S., Kruspe, A., and Zhu, X.~X.
\newblock An advanced dirichlet prior network for out-of-distribution detection
  in remote sensing.
\newblock \emph{IEEE Transactions on Geoscience and Remote Sensing}, 2022.

\bibitem[Geng(2016)]{geng2016label}
Geng, X.
\newblock Label distribution learning.
\newblock \emph{IEEE Transactions on Knowledge and Data Engineering},
  28\penalty0 (7):\penalty0 1734--1748, 2016.

\bibitem[Geng et~al.(2013)Geng, Yin, and Zhou]{geng2013facial}
Geng, X., Yin, C., and Zhou, Z.-H.
\newblock Facial age estimation by learning from label distributions.
\newblock \emph{IEEE transactions on pattern analysis and machine
  intelligence}, 35\penalty0 (10):\penalty0 2401--2412, 2013.

\bibitem[Guo et~al.(2017)Guo, Pleiss, Sun, and Weinberger]{guo2017calibration}
Guo, C., Pleiss, G., Sun, Y., and Weinberger, K.~Q.
\newblock On calibration of modern neural networks.
\newblock In \emph{International Conference on Machine Learning}, pp.\
  1321--1330. PMLR, 2017.

\bibitem[Hinton et~al.(2015)Hinton, Vinyals, and Dean]{hinton2015distilling}
Hinton, G., Vinyals, O., and Dean, J.
\newblock Distilling the knowledge in a neural network.
\newblock \emph{arXiv preprint arXiv:1503.02531}, 2015.

\bibitem[Hollemans(2020)]{reliabilitydiagram}
Hollemans, M.
\newblock Reliability diagrams.
\newblock \url{https://github.com/hollance/reliability-diagrams}, 2020.
\newblock [Online, Accessed: 2022-01-10].

\bibitem[Hu et~al.(2018)Hu, Ghamisi, and Zhu]{hu2018feature}
Hu, J., Ghamisi, P., and Zhu, X.~X.
\newblock Feature extraction and selection of sentinel-1 dual-pol data for
  global-scale local climate zone classification.
\newblock \emph{ISPRS International Journal of Geo-Information}, 7\penalty0
  (9):\penalty0 379, 2018.

\bibitem[H{\"u}llermeier \& Waegeman(2021)H{\"u}llermeier and
  Waegeman]{hullermeier2021aleatoric}
H{\"u}llermeier, E. and Waegeman, W.
\newblock Aleatoric and epistemic uncertainty in machine learning: An
  introduction to concepts and methods.
\newblock \emph{Machine Learning}, 110\penalty0 (3):\penalty0 457--506, 2021.

\bibitem[Lakshminarayanan et~al.(2017)Lakshminarayanan, Pritzel, and
  Blundell]{lakshminarayanan2017simple}
Lakshminarayanan, B., Pritzel, A., and Blundell, C.
\newblock Simple and scalable predictive uncertainty estimation using deep
  ensembles.
\newblock \emph{Advances in Neural Information Processing Systems}, 30, 2017.

\bibitem[Lampert et~al.(2013)Lampert, Nickisch, and
  Harmeling]{lampert2013attribute}
Lampert, C.~H., Nickisch, H., and Harmeling, S.
\newblock Attribute-based classification for zero-shot visual object
  categorization.
\newblock \emph{IEEE transactions on pattern analysis and machine
  intelligence}, 36\penalty0 (3):\penalty0 453--465, 2013.

\bibitem[Lelovics et~al.(2014)Lelovics, Unger, G{\'a}l, and
  G{\'a}l]{lelovics2014design}
Lelovics, E., Unger, J., G{\'a}l, T., and G{\'a}l, C.~V.
\newblock Design of an urban monitoring network based on local climate zone
  mapping and temperature pattern modelling.
\newblock \emph{Climate research}, 60\penalty0 (1):\penalty0 51--62, 2014.

\bibitem[Lukasik et~al.(2020)Lukasik, Bhojanapalli, Menon, and
  Kumar]{lukasik2020does}
Lukasik, M., Bhojanapalli, S., Menon, A., and Kumar, S.
\newblock Does label smoothing mitigate label noise?
\newblock In \emph{International Conference on Machine Learning}, pp.\
  6448--6458. PMLR, 2020.

\bibitem[Luo et~al.(2021)Luo, Wang, Ou, He, and Li]{luo2021neighbor}
Luo, J., Wang, Y., Ou, Y., He, B., and Li, B.
\newblock Neighbor-based label distribution learning to model label ambiguity
  for aerial scene classification.
\newblock \emph{Remote Sensing}, 13\penalty0 (4):\penalty0 755, 2021.

\bibitem[Maddox et~al.(2019)Maddox, Izmailov, Garipov, Vetrov, and
  Wilson]{maddox2019simple}
Maddox, W.~J., Izmailov, P., Garipov, T., Vetrov, D.~P., and Wilson, A.~G.
\newblock A simple baseline for bayesian uncertainty in deep learning.
\newblock \emph{Advances in Neural Information Processing Systems},
  32:\penalty0 13153--13164, 2019.

\bibitem[Malinin \& Gales(2018)Malinin and Gales]{malinin2018predictive}
Malinin, A. and Gales, M.
\newblock Predictive uncertainty estimation via prior networks.
\newblock In \emph{Proceedings of the 32nd International Conference on Neural
  Information Processing Systems}, pp.\  7047--7058, 2018.

\bibitem[Mills et~al.(2015)Mills, Ching, See, Bechtel, and
  Foley]{mills2015introduction}
Mills, G., Ching, J., See, L., Bechtel, B., and Foley, M.
\newblock An introduction to the wudapt project.
\newblock In \emph{Proceedings of the 9th International Conference on Urban
  Climate, Toulouse, France}, pp.\  20--24, 2015.

\bibitem[Mozafari et~al.(2018)Mozafari, Gomes, Le{\~a}o, Janny, and
  Gagn{\'e}]{mozafari2018attended}
Mozafari, A.~S., Gomes, H.~S., Le{\~a}o, W., Janny, S., and Gagn{\'e}, C.
\newblock Attended temperature scaling: a practical approach for calibrating
  deep neural networks.
\newblock \emph{arXiv preprint arXiv:1810.11586}, 2018.

\bibitem[M{\"u}ller et~al.(2019)M{\"u}ller, Kornblith, and
  Hinton]{muller2019does}
M{\"u}ller, R., Kornblith, S., and Hinton, G.
\newblock When does label smoothing help?
\newblock \emph{arXiv preprint arXiv:1906.02629}, 2019.

\bibitem[Niculescu-Mizil \& Caruana(2005)Niculescu-Mizil and
  Caruana]{niculescu2005predicting}
Niculescu-Mizil, A. and Caruana, R.
\newblock Predicting good probabilities with supervised learning.
\newblock In \emph{Proceedings of the 22nd international conference on Machine
  learning}, pp.\  625--632, 2005.

\bibitem[Nixon et~al.(2019)Nixon, Dusenberry, Zhang, Jerfel, and
  Tran]{nixon2019measuring}
Nixon, J., Dusenberry, M.~W., Zhang, L., Jerfel, G., and Tran, D.
\newblock Measuring calibration in deep learning.
\newblock In \emph{CVPR Workshops}, volume~2, 2019.

\bibitem[Peng et~al.(2006)Peng, Radivojac, Vucetic, Dunker, and
  Obradovic]{peng2006length}
Peng, K., Radivojac, P., Vucetic, S., Dunker, A.~K., and Obradovic, Z.
\newblock Length-dependent prediction of protein intrinsic disorder.
\newblock \emph{BMC bioinformatics}, 7\penalty0 (1):\penalty0 1--17, 2006.

\bibitem[Perera \& Emmanuel(2018)Perera and Emmanuel]{perera2018local}
Perera, N. and Emmanuel, R.
\newblock A “local climate zone” based approach to urban planning in
  colombo, sri lanka.
\newblock \emph{Urban Climate}, 23:\penalty0 188--203, 2018.

\bibitem[Peterson et~al.(2019)Peterson, Battleday, Griffiths, and
  Russakovsky]{peterson2019human}
Peterson, J.~C., Battleday, R.~M., Griffiths, T.~L., and Russakovsky, O.
\newblock Human uncertainty makes classification more robust.
\newblock In \emph{Proceedings of the IEEE/CVF International Conference on
  Computer Vision}, pp.\  9617--9626, 2019.

\bibitem[Platt et~al.(1999)]{platt1999probabilistic}
Platt, J. et~al.
\newblock Probabilistic outputs for support vector machines and comparisons to
  regularized likelihood methods.
\newblock \emph{Advances in large margin classifiers}, 10\penalty0
  (3):\penalty0 61--74, 1999.

\bibitem[Qiu et~al.(2019)Qiu, Mou, Schmitt, and Zhu]{qiu2019local}
Qiu, C., Mou, L., Schmitt, M., and Zhu, X.~X.
\newblock Local climate zone-based urban land cover classification from
  multi-seasonal sentinel-2 images with a recurrent residual network.
\newblock \emph{ISPRS Journal of Photogrammetry and Remote Sensing},
  154:\penalty0 151--162, 2019.

\bibitem[Qiu et~al.(2020{\natexlab{a}})Qiu, Mou, Schmitt, and
  Zhu]{qiu2020fusing}
Qiu, C., Mou, L., Schmitt, M., and Zhu, X.~X.
\newblock Fusing multiseasonal sentinel-2 imagery for urban land cover
  classification with multibranch residual convolutional neural networks.
\newblock \emph{IEEE Geoscience and Remote Sensing Letters}, 17\penalty0
  (10):\penalty0 1787--1791, 2020{\natexlab{a}}.

\bibitem[Qiu et~al.(2020{\natexlab{b}})Qiu, Tong, Schmitt, Bechtel, and
  Zhu]{qiu2020multilevel}
Qiu, C., Tong, X., Schmitt, M., Bechtel, B., and Zhu, X.~X.
\newblock Multilevel feature fusion-based cnn for local climate zone
  classification from sentinel-2 images: Benchmark results on the so2sat lcz42
  dataset.
\newblock \emph{IEEE Journal of Selected Topics in Applied Earth Observations
  and Remote Sensing}, 13:\penalty0 2793--2806, 2020{\natexlab{b}}.

\bibitem[Quanz et~al.(2018)Quanz, Ulrich, Fenner, Holtmann, and
  Eimermacher]{quanz2018micro}
Quanz, J.~A., Ulrich, S., Fenner, D., Holtmann, A., and Eimermacher, J.
\newblock Micro-scale variability of air temperature within a local climate
  zone in berlin, germany, during summer.
\newblock \emph{Climate}, 6\penalty0 (1):\penalty0 5, 2018.

\bibitem[Rolnick et~al.(2017)Rolnick, Veit, Belongie, and
  Shavit]{rolnick2017deep}
Rolnick, D., Veit, A., Belongie, S., and Shavit, N.
\newblock Deep learning is robust to massive label noise.
\newblock \emph{arXiv preprint arXiv:1705.10694}, 2017.

\bibitem[Russwurm et~al.(2020)Russwurm, Ali, Zhu, Gal, and
  Korner]{russwurm2020model}
Russwurm, M., Ali, M., Zhu, X.~X., Gal, Y., and Korner, M.
\newblock Model and data uncertainty for satellite time series forecasting with
  deep recurrent models.
\newblock In \emph{IGARSS 2020-2020 IEEE International Geoscience and Remote
  Sensing Symposium}, 2020.

\bibitem[Samsonov \& Trigub(2017)Samsonov and Trigub]{samsonov2017towards}
Samsonov, T. and Trigub, K.
\newblock Towards computation of urban local climate zones (lcz) from
  openstreetmap data.
\newblock In \emph{Proceedings of the 14th International Conference on
  GeoComputation, Leeds, UK}, pp.\  4--7, 2017.

\bibitem[Sensoy et~al.(2018)Sensoy, Kaplan, and Kandemir]{sensoy2018evidential}
Sensoy, M., Kaplan, L., and Kandemir, M.
\newblock Evidential deep learning to quantify classification uncertainty.
\newblock In \emph{Proceedings of the 32nd International Conference on Neural
  Information Processing Systems}, pp.\  3183--3193, 2018.

\bibitem[Stewart(2011)]{stewart2011local}
Stewart, I.
\newblock Local climate zones: Origins, development, and application to urban
  heat island studies.
\newblock In \emph{Proceedings of the Annual Meeting of the American
  Association of Geographers, Seattle, WA, USA}, pp.\  12--16, 2011.

\bibitem[Stewart \& Oke(2012)Stewart and Oke]{stewart2012local}
Stewart, I.~D. and Oke, T.~R.
\newblock Local climate zones for urban temperature studies.
\newblock \emph{Bulletin of the American Meteorological Society}, 93\penalty0
  (12):\penalty0 1879--1900, 2012.

\bibitem[Stewart et~al.(2014)Stewart, Oke, and
  Krayenhoff]{stewart2014evaluation}
Stewart, I.~D., Oke, T.~R., and Krayenhoff, E.~S.
\newblock Evaluation of the ‘local climate zone’scheme using temperature
  observations and model simulations.
\newblock \emph{International journal of climatology}, 34\penalty0
  (4):\penalty0 1062--1080, 2014.

\bibitem[Szegedy et~al.(2016)Szegedy, Vanhoucke, Ioffe, Shlens, and
  Wojna]{szegedy2016rethinking}
Szegedy, C., Vanhoucke, V., Ioffe, S., Shlens, J., and Wojna, Z.
\newblock Rethinking the inception architecture for computer vision.
\newblock In \emph{Proceedings of the IEEE conference on computer vision and
  pattern recognition}, pp.\  2818--2826, 2016.

\bibitem[Thomas et~al.(2014)Thomas, Sherin, Ansar, and
  Zachariah]{thomas2014analysis}
Thomas, G., Sherin, A., Ansar, S., and Zachariah, E.
\newblock Analysis of urban heat island in kochi, india, using a modified local
  climate zone classification.
\newblock \emph{Procedia Environmental Sciences}, 21:\penalty0 3--13, 2014.

\bibitem[Verdonck et~al.(2017)Verdonck, Okujeni, van~der Linden, Demuzere,
  De~Wulf, and Van~Coillie]{verdonck2017influence}
Verdonck, M.-L., Okujeni, A., van~der Linden, S., Demuzere, M., De~Wulf, R.,
  and Van~Coillie, F.
\newblock Influence of neighbourhood information on ‘local climate
  zone’mapping in heterogeneous cities.
\newblock \emph{International Journal of Applied Earth Observation and
  Geoinformation}, 62:\penalty0 102--113, 2017.

\bibitem[Wang et~al.(2017)Wang, Wu, Zhang, Zhang, and Li]{wang2017unsupervised}
Wang, F., Wu, Y., Zhang, P., Zhang, Q., and Li, M.
\newblock Unsupervised sar image segmentation using ambiguity label information
  fusion in triplet markov fields model.
\newblock \emph{IEEE Geoscience and Remote Sensing Letters}, 14\penalty0
  (9):\penalty0 1479--1483, 2017.

\bibitem[Yang et~al.(2019)Yang, Jin, Xiao, Jin, Xia, Li, and
  Wang]{yang2019local}
Yang, J., Jin, S., Xiao, X., Jin, C., Xia, J.~C., Li, X., and Wang, S.
\newblock Local climate zone ventilation and urban land surface temperatures:
  Towards a performance-based and wind-sensitive planning proposal in
  megacities.
\newblock \emph{Sustainable Cities and Society}, 47:\penalty0 101487, 2019.

\bibitem[Yoo et~al.(2019)Yoo, Han, Im, and Bechtel]{yoo2019comparison}
Yoo, C., Han, D., Im, J., and Bechtel, B.
\newblock Comparison between convolutional neural networks and random forest
  for local climate zone classification in mega urban areas using landsat
  images.
\newblock \emph{ISPRS Journal of Photogrammetry and Remote Sensing},
  157:\penalty0 155--170, 2019.

\bibitem[Yuan et~al.(2020)Yuan, Tay, Li, Wang, and Feng]{yuan2020revisiting}
Yuan, L., Tay, F.~E., Li, G., Wang, T., and Feng, J.
\newblock Revisiting knowledge distillation via label smoothing regularization.
\newblock In \emph{Proceedings of the IEEE/CVF Conference on Computer Vision
  and Pattern Recognition}, pp.\  3903--3911, 2020.

\bibitem[Zhang et~al.(2021)Zhang, Jiang, Hou, Wei, Han, Li, and
  Cheng]{zhang2021delving}
Zhang, C.-B., Jiang, P.-T., Hou, Q., Wei, Y., Han, Q., Li, Z., and Cheng, M.-M.
\newblock Delving deep into label smoothing.
\newblock \emph{IEEE Transactions on Image Processing}, 30:\penalty0
  5984--5996, 2021.

\bibitem[Zhou et~al.(2021)Zhou, Greenspan, Davatzikos, Duncan, Van~Ginneken,
  Madabhushi, Prince, Rueckert, and Summers]{zhou2021review}
Zhou, S.~K., Greenspan, H., Davatzikos, C., Duncan, J.~S., Van~Ginneken, B.,
  Madabhushi, A., Prince, J.~L., Rueckert, D., and Summers, R.~M.
\newblock A review of deep learning in medical imaging: Imaging traits,
  technology trends, case studies with progress highlights, and future
  promises.
\newblock \emph{Proceedings of the IEEE}, 2021.

\bibitem[Zhu et~al.(2020)Zhu, Hu, Qiu, Shi, Kang, Mou, Bagheri, Haberle, Hua,
  Huang, et~al.]{zhu2020so2sat}
Zhu, X.~X., Hu, J., Qiu, C., Shi, Y., Kang, J., Mou, L., Bagheri, H., Haberle,
  M., Hua, Y., Huang, R., et~al.
\newblock So2sat lcz42: a benchmark data set for the classification of global
  local climate zones [software and data sets].
\newblock \emph{IEEE Geoscience and Remote Sensing Magazine}, 8\penalty0
  (3):\penalty0 76--89, 2020.

\bibitem[Zhu et~al.(2022)Zhu, Qiu, Hu, Shi, Wang, Schmitt, and
  Taubenböck]{zhu2022}
Zhu, X.~X., Qiu, C., Hu, J., Shi, Y., Wang, Y., Schmitt, M., and Taubenböck,
  H.
\newblock The urban morphology on our planet – global perspectives from
  space.
\newblock \emph{Remote Sensing of Environment}, 269:\penalty0 112794, 2022.

\end{thebibliography}
\bibliographystyle{icml2022}



\newpage

\appendix

\section{Appendix}

\subsection{Data Split}
\label{appendix:split}

As part of the data labeling performed by \citet{zhu2020so2sat}, a labeling confidence study was included. For this study, 10 individual human annotators were asked to cast a label vote for a subset of cities which were included in the original So2Sat LCZ42 data set. In particular, the cities were subsetted to be European cities, which the labelers were assumed to be confident in labeling. Overall, a confidence of $85\%$ was stated by the authors. Next to European cities, additional urban conglomerates from less developed regions around the globe were added. This additional data ensures a stronger class balance, as some classes such as class 7 (Lightweight low-rise buildings) are underrepresented in European cities. The entire list of cities, as well as addon areas for which the label evaluation was performed, is given in Table \ref{table:cities_list}.  

\begin{table*}[h!]
    \caption{Cities and addon areas in the evaluation subset of the So2Sat LCZ42 data set. }
    \vskip 0.15in
    \centering
    \begin{tabular}{llcll}
     \toprule \multicolumn{2}{c}{Cities} & & \multicolumn{2}{c}{Addon Areas} \\
     \cmidrule{1-2} \cmidrule{4-5} 
     Amsterdam & Berlin & & Guangzhou & Islamabad \\
     Cologne & London & & Jakarta & Los Angeles \\
     Madrid & Milan & & Moscow & Mumbai \\
     Munich & Paris & & Nairobi & Riodejaneiro \\
     Rome & Zurich & & & \\
     \bottomrule
     \end{tabular}
     \label{table:cities_list}
\end{table*}

In order to geographically separate training and testing data, the corresponding data sets were formed by mutually exclusive subsets of the above cities and addon areas. The split on a city level was specifically chosen not to avoid learning the overall similar data distribution, as all images come from European cities, but rather to train on one set of cities and predict as well as evaluate the method on another set of cities. In Table \ref{table:cities_list_split}, the cities in the respective data sets are listed. Note that for the separation of validation and testing data, a random split of the cities and addon areas was performed, halving the entirety of data points into validation and testing data. 

\begin{table*}[h!]
    \caption{Manual split of the cities and addon areas of the evaluation set into training data and non-training data, which was in a second step split randomly by half into validation and testing data. }
    \vskip 0.15in
    \centering
    \begin{tabular}{lcllcll}
     \toprule & & Cities & & & Addon Areas & \\
     \cmidrule{3-4} \cmidrule{6-7} 
     \multirow{3}{*}{Training Data} & & Amsterdam & Berlin & & Guangzhou & Islamabad \\
     & & Cologne & London & & Jakarta & Nairobi \\
     & & Milan & Rome & & & \\
     \cmidrule{1-1} \cmidrule{3-4} \cmidrule{6-7}
     \multirow{2}{*}{Validation / Test Data} & & Madrid & Munich & & Los Angeles & Moscow \\
     & & Paris & Zurich & & Mumbai & Riodejaneiro \\
     \bottomrule
    \end{tabular}
    \label{table:cities_list_split}
    \vskip 0.15in
\end{table*}

Regarding the choice of cities, next to the geographical separation a strong emphasis was laid on balancing the occurring class frequencies in between the data sets. For the investigated urban classes (LCZs 1-10), the class frequencies of the different data sets relative to the number of samples in the entire data set can be found in Table \ref{table:class_freqs}. Note that for the ground-truth label, here the majority vote $\bm{y}_{\text{max}}$ was taken. Images with incomplete label distributions, that is, those having a majority vote from the urban classes and one or more individual votes from the non-urban classes occur very rarely ($\sim 0.1 \%)$. The outlier votes have been excluded from the analysis. As can be deduced from Table \ref{table:class_freqs}, the classes are not perfectly balanced among the training, validation and test data sets. Yet a significant effort was spent on finding a good split on city level that still retains a moderate level of class imbalance. This imbalance among the data sets adds complexity to the imbalance among the different classes already present in the data.  

\begin{table*}[h!]
    \caption{Relative class frequencies of the urban classes within the training, validation and testing set. Totals are listed in the right and bottom entries. }
    \vskip 0.15in
    \centering
    \small
    \begin{tabular}{lccccccccccccr}
         \toprule \addlinespace & & \multicolumn{10}{c}{Urban LCZ} & &  \\ \addlinespace
         & & 1 & 2 & 3 & 4 & 5 & 6 & 7 & 8 & 9 & 10 & & $\sum$ \\ \addlinespace
         \cmidrule{3-12} \cmidrule{14-14} \addlinespace Training Data & & 0.83 & 0.56 & 0.87 & 0.61 & 0.83 & 0.74 & 0.65 & 0.74 & 0.65 & 0.94 & & 43k \\ \addlinespace
         Validation Data & & 0.09 & 0.22 & 0.07 & 0.20 & 0.08 & 0.13 & 0.18 & 0.13 & 0.17 & 0.03 & & 8k \\ \addlinespace
         Test Data & & 0.09 & 0.22 & 0.07 & 0.20 & 0.08 & 0.13 & 0.18 & 0.13 & 0.17 & 0.03 & & 8k \\ \addlinespace
         \cmidrule{3-12} \cmidrule{14-14} \addlinespace $\sum$  & & $<$ 1k & 10k & 2k & 1k & 11k & 18k & 3k & 9k & 1k & 3k & & 59k \\ \addlinespace
         \bottomrule
    \end{tabular}
    \label{table:class_freqs}
    \vskip 0.15in
\end{table*}

\subsection{Human Label Uncertainty}
\label{appendix:human}

\begin{figure*}[h!]
    \vskip 0.1in
    \subfloat[]{
    \includegraphics[clip, width=0.45\textwidth]{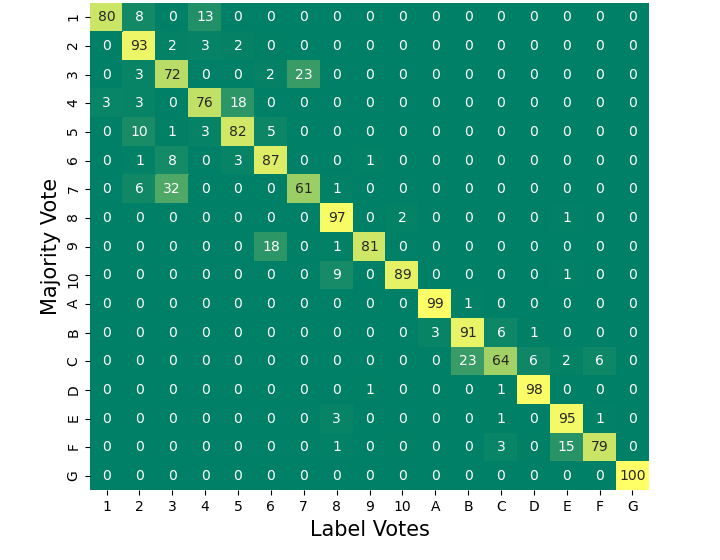}
    }
    \qquad
    \subfloat[]{
    \includegraphics[clip, width=0.45\textwidth]{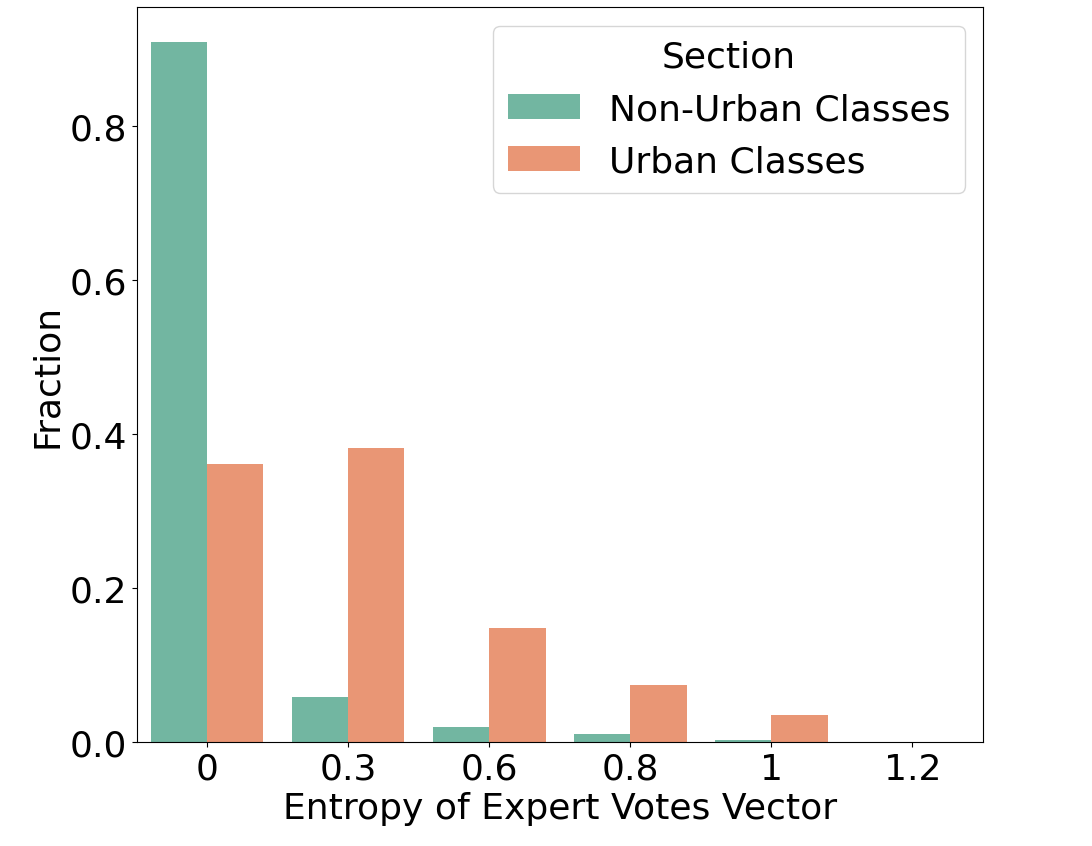}
    }
    \caption{(a) Confusion matrix of label votes for the evaluation data set; (b) Entropies of the voting distributions. }
    \label{fig:confusion}
\end{figure*}

The 10 label votes received from each of the ten independent remote sensing experts inherently store a notion of label uncertainty. This uncertainty is visualized in Figure \ref{fig:confusion} by means of a confusion matrix between the individual label votes $ V_1^{(i)},\dots,V_J^{(i)}$ and the majority vote $\bm{y}^{(i)}_{\text{max}}$ for all images $i=1,\dots,n$, as well as a plot showing the entropies of the individual label distributions. In detail, for the given label distribution $\bm{y}^{(i)}_{\text{distr}}$ (see  (\ref{eq:distr})), we compute the information theoretic (Shannon) entropy of the distribution given for image $i$ via \begin{equation*}
    H(\bm{y}^{(i)}_{\text{distr}}) = -\displaystyle\sum_{k=1}^K \bm{y}^{(i)}_{\text{distr}, k} \log \bm{y}^{(i)}_{\text{distr}, k}
\end{equation*}
and plot the resulting values in a bar plot grouped by the affiliation of the respective majority vote $\bm{y}^{(i)}_{\text{max}}$ (see (\ref{eq:majority})) to either the urban or non-urban classes. Whereas zero entropy occurs when all voters agree on a label for an image, maximum entropy is achieved for a uniform distribution across all labels (which cannot occur here because the number of classes exceeds the number of voters). Clearly visible is the higher average entropy of the vote vectors for the urban classes, which corresponds to a higher uncertainty associated with the respective satellite images. We chose to limit our analysis to the urban classes due to the large share of images with the majority vote belonging to non-urban classes that at the same time have zero entropy in the votes.   

\subsection{Methodological}
\label{appendix:method}

\subsubsection{Calibration and Generalization Metrics}

Next to the ECE earlier derived, further metrics for evaluating the calibration performance of a classification model exist and have been reported in the paper. A simpler version of the ECE, which only considers the maximum gap between confidence and accuracy out of all bins considered, is the Maximum Calibration Error (MCE). Given the predictions $\bm{\hat{y}}^{(i)}\,,\, i=1,\dots,n$ and corresponding confidences $\hat{p}^{(i)}\,,\, i=1,\dots,n$, the computation is as follows: 

\begin{equation*}
     \text{MCE} = \max\limits_m |\,\text{acc}(I_m) -  \text{conf}(I_m)\,|
\end{equation*}

While the MCE gives a first indication whether the evaluated classifier is severely miscalibrated for a certain bin, the same downside appears as for the ECE, namely that the included metrics are not considered class-specific. For this purpose \citet{nixon2019measuring} introduced the Static Calibration Error (SCE), which measures the accuracies and confidences on a class level. For class $k$ and within bin $m$, these are depicted by $\text{acc}(m,k)$ and $\text{conf}(m,k)$, leading to the formula for the SEC given by

\begin{equation*}
    \text{SCE} = \frac{1}{K} \displaystyle\sum_{k=1}^K\displaystyle\sum_{m=1}^M \frac{n_{mk}}{n} | \text{acc}(m,k) - \text{conf}(m,k) | 
\end{equation*}

where $n_{bk}$ denotes the number of samples belonging to class $k$ (with respect to the majority vote) within bin $b$. Not only is the full information of the predictive distribution used instead of having to rely solely on the highest predicted probability, but the class-specific scores are also weighted with respect to the class share within the respective bin. Nonetheless, the dependency of all introduced calibration metrics on the chosen number of bins remains. To get a clearer picture, more results based on the same trained models as in the paper, but with differing number of bins, are reported in Section \ref{appendix:ablation}. 

The measures of generalization as shown in Table \ref{table:ce20} namely are the cross-entropies between the predictive distribution of the network $p_{\theta}(y|x)$ and the one-hot encoded label based on the majority vote $y_{\text{max}}$ or the distributional label $y_{\text{distr}}$. They are termed "CE One-hot" and "CE Distr" and given for image $i$ via 

\begin{align*}
    \text{CE}(y_{\text{max}}^{(i)} , p_{\theta}(y^{(i)}|x)) &= -\displaystyle\sum_{k=1}^K y_{\text{max}, k}^{(i)} \log(p_{\theta}(y^{(i)}=k|x)  \,\, \text{and} \\
    \text{CE}(y_{\text{distr}}^{(i)} , p_{\theta}(y^{(i)}|x)) &= -\displaystyle\sum_{k=1}^K y_{\text{distr}, k}^{(i)} \log(p_{\theta}(y^{(i)}=k|x) \, \text{, respectively.  }
\end{align*}

\subsubsection{Calibration Techniques}

Presented as a straightforward and effective calibration strategy, \citet{guo2017calibration} introduced Temperature scaling. The technique can be sen as a simplification to Platt scaling \citep{platt1999probabilistic} which scales the logits predicted by the classifier by a constant parameter. So given the predicted logit of the neural classifier $f_{\theta}(x)$ for image $i$ via $z^{(i)}$, the corresponding softmax prediction of the scaled logit is given via $$\text{softmax}(z^{(i)}) = \frac{\exp(z^{(i)} / \, T)}{\sum_k \exp(z^{(i)}_k / \, T)} = \Big(\frac{\exp(z^{(i)}_1 / \, T)}{\sum_k \exp(z^{(i)}_k / \, T)},\dots,\frac{\exp(z^{(i)}_K / \, T)}{\sum_k \exp(z^{(i)}_k / \, T)}\Big)$$ 

As described by \citet{guo2017calibration}, we optimized the parameter T (termed temperature) with respect to the negative log-likelihood on the validation data set. Adding to that method, Label smoothing represents another quick and easy to implement off-the-shelf calibration method. As opposed to temperature scaling, the method is not applied post-hoc, but uses a hyperparameter $\alpha$ to scale the labels before training. Given the label $y^{(i)}$ for image $i$, the scaled version is then received via $$y^{(i)}_{\text{smoothed}} = \alpha \cdot u_K + (1 - \alpha) \cdot y^{(i)}$$ where $u_K$ denotes the uniform distribution over the $K$ classes. The hyperparameter is not directly optimized, but empirically chosen. As a third calibration option, we utilized Monte Carlo Dropout \citep{gal2016dropout}. Given the already trained networks, we left the dropout layers of Sen2LCZ active during prediction and by doing so created a set of 20 unique predictions. After averaging the softmax vectors of the individual predictions, we proceeded with deriving the calibration and generalization metrics as before. 

\subsection{Further Analyses}
\label{appendix:ablation}

The derivation of the calibration errors depends on the chosen number of bins. In order to further investigate the results presented in the paper, we present the calibration metrics for different numbers of bins in the following. The binning has been performed using 10, 15 and 25 equally-sized bins for which the results are presented in Tables \ref{table:ce10}, \ref{table:ce15} and \ref{table:ce25}, respectively. 

\newpage

\begin{table}[h!]
    \small
    \caption{Cross-Entropies between predicted softmax probabilities and labels on the test set as well as calibration errors, averaged over five runs. CE = Cross entropy, LS = Label smoothing, TS = Temperature Scaling, MC-Drop = Monte Carlo Dropout. Binning was performed using 10 equally-sized bins. }
    \centering
    \begin{tabular}{lccccc}
    \toprule \addlinespace
    & CE One-hot $\downarrow$ & CE Distr. $\downarrow$ & ECE $\downarrow$ & MCE $\downarrow$ & SCE $\downarrow$ \\ \addlinespace
    \cmidrule{2-6} 
    One-hot & 1.12 $\pm$ 0.05 & 1.38 $\pm$ 0.07 & 9.71 $\pm$ 3.18 & 21.31 $\pm$ 5.21 & 1.01 $\pm$ 0.50 \\ 
    + LS & 1.05 $\pm$ 0.01 & 1.23 $\pm$ 0.03 & 6.92 $\pm$ 2.98 & 17.84 $\pm$ 4.13 & 1.09 $\pm$ 0.25 \\ 
    + TS & 1.00 $\pm$ 0.13 & 1.17 $\pm$ 0.07 & 3.95 $\pm$ 2.49 & 12.43 $\pm$ 7.90 & 1.42 $\pm$ 0.23 \\ 
    + LS \& TS & 1.02 $\pm$ 0.03 & 1.18 $\pm$ 0.02 & 2.82 $\pm$ 1.16 & 10.77 $\pm$ 4.49 & 1.30 $\pm$ 0.03 \\
    + MC-Drop & 1.12 $\pm$ 0.05 & 1.37 $\pm$ 0.06 & 9.43 $\pm$ 3.10 & 20.23 $\pm$ 5.15 & 1.02 $\pm$ 0.50 \\ 
    + LS \& MC-Drop & 1.05 $\pm$ 0.01 & 1.23 $\pm$ 0.03 & 6.56 $\pm$ 2.86 & 14.36 $\pm$ 5.74 & 1.11 $\pm$ 0.25 \\ \addlinespace 
    Distr. & 1.06 $\pm$ 0.07 & 1.21 $\pm$ 0.07 & 5.44 $\pm$ 1.46 & 13.91 $\pm$ 4.05 & 1.19 $\pm$ 0.20 \\ 
    + LS & 0.98 $\pm$ 0.03 & 1.08 $\pm$ 0.02 & 8.20 $\pm$ 2.56 & 13.37 $\pm$ 4.47 & 1.71 $\pm$ 0.06 \\  
    + TS & 0.96 $\pm$ 0.09 & 1.07 $\pm$ 0.07 & 5.60 $\pm$ 2.51 & 14.99 $\pm$ 4.05 & 1.71 $\pm$ 0.15 \\ 
    + LS \& TS & 0.95 $\pm$ 0.04 & 1.05 $\pm$ 0.04 & 4.03 $\pm$ 1.52 & 14.78 $\pm$ 2.37 & 1.57 $\pm$ 0.10 \\ \addlinespace 
    \bottomrule
    \end{tabular}
    \label{table:ce10}
\end{table}

\vspace{-0.5cm}

\begin{table}[h!]
    \small
    \caption{Cross-Entropies between predicted softmax probabilities and labels on the test set as well as calibration errors, averaged over five runs. CE = Cross entropy, LS = Label smoothing, TS = Temperature Scaling, MC-Drop = Monte Carlo Dropout. Binning was performed using 15 equally-sized bins. }
    \centering
    \begin{tabular}{lccccc}
    \toprule \addlinespace
    & CE One-hot $\downarrow$ & CE Distr. $\downarrow$ & ECE $\downarrow$ & MCE $\downarrow$ & SCE $\downarrow$ \\ \addlinespace
    \cmidrule{2-6} 
    One-hot & 1.12 $\pm$ 0.05 & 1.38 $\pm$ 0.07 & 9.54 $\pm$ 3.25 & 23.05 $\pm$ 5.35 & 1.01 $\pm$ 0.51 \\  
    + LS & 1.05 $\pm$ 0.01 & 1.23 $\pm$ 0.03 & 7.35 $\pm$ 2.69 & 19.40 $\pm$ 5.06 & 1.10 $\pm$ 0.25 \\ 
    + TS & 1.00 $\pm$ 0.13 & 1.17 $\pm$ 0.07 & 4.07 $\pm$ 2.41 & 14.51 $\pm$ 10.23 & 1.43 $\pm$ 0.23 \\ 
    + LS \& TS & 1.02 $\pm$ 0.03 & 1.18 $\pm$ 0.02 & 3.08 $\pm$ 0.99 & 10.76 $\pm$ 4.53 & 1.31 $\pm$ 0.03 \\
    + MC-Drop & 1.12 $\pm$ 0.05 & 1.37 $\pm$ 0.06 & 9.28 $\pm$ 3.17 & 32.88 $\pm$ 26.97 & 1.01 $\pm$ 0.51 \\ 
    + LS \& MC-Drop & 1.05 $\pm$ 0.01 & 1.23 $\pm$ 0.03 & 7.11 $\pm$ 2.54 & 33.41 $\pm$ 26.56 & 1.11 $\pm$ 0.25 \\ \addlinespace 
    Distr. & 1.06 $\pm$ 0.07 & 1.21 $\pm$ 0.07 & 5.76 $\pm$ 1.18 & 14.76 $\pm$ 3.81 & 1.19 $\pm$ 0.19 \\ 
    + LS & 0.98 $\pm$ 0.03 & 1.08 $\pm$ 0.02 & 8.16 $\pm$ 2.69 & 15.30 $\pm$ 3.20 & 1.72 $\pm$ 0.06 \\ 
    + TS & 0.96 $\pm$ 0.09 & 1.07 $\pm$ 0.07 & 5.87 $\pm$ 2.61 & 15.23 $\pm$ 3.96 & 1.71 $\pm$ 0.15 \\ 
    + LS \& TS & 0.95 $\pm$ 0.04 & 1.05 $\pm$ 0.04 & 4.09 $\pm$ 1.42 & 14.79 $\pm$ 2.43 & 1.57 $\pm$ 0.10 \\ \addlinespace 
    \bottomrule
    \end{tabular}
    \label{table:ce15}
\end{table}

\vspace{-0.5cm}

\begin{table}[h!]
    \small
    \caption{Cross-Entropies between predicted softmax probabilities and labels on the test set as well as calibration errors, averaged over five runs. CE = Cross entropy, LS = Label smoothing, TS = Temperature Scaling, MC-Drop = Monte Carlo Dropout. Binning was performed using 25 equally-sized bins. }
    \centering
    \begin{tabular}{lccccc}
    \toprule \addlinespace
    & CE One-hot $\downarrow$ & CE Distr. $\downarrow$ & ECE $\downarrow$ & MCE $\downarrow$ & SCE $\downarrow$ \\ \addlinespace
    \cmidrule{2-6} 
    One-hot & 1.12 $\pm$ 0.05 & 1.38 $\pm$ 0.07 & 9.83 $\pm$ 2.88 & 27.50 $\pm$ 7.49 & 1.03 $\pm$ 0.51 \\ 
    + LS & 1.05 $\pm$ 0.01 & 1.23 $\pm$ 0.03 & 7.22 $\pm$ 2.98 & 22.50 $\pm$ 2.77 & 1.12 $\pm$ 0.24 \\ 
    + TS & 1.00 $\pm$ 0.13 & 1.17 $\pm$ 0.07 & 4.23 $\pm$ 2.37 & 16.69 $\pm$ 10.43 & 1.44 $\pm$ 0.23 \\ 
    + LS \& TS & 1.02 $\pm$ 0.03 & 1.18 $\pm$ 0.02 & 3.20 $\pm$ 0.84 & 13.86 $\pm$ 5.32 & 1.32 $\pm$ 0.03 \\
    + MC-Drop & 1.12 $\pm$ 0.05 & 1.37 $\pm$ 0.06 & 9.57 $\pm$ 2.92 & 33.34 $\pm$ 26.43 & 1.04 $\pm$ 0.50 \\
    + LS \& MC-Drop & 1.05 $\pm$ 0.01 & 1.23 $\pm$ 0.03 & 7.00 $\pm$ 2.77 & 22.57 $\pm$ 3.13 & 1.12 $\pm$ 0.24 \\ \addlinespace 
    Distr. & 1.06 $\pm$ 0.07 & 1.21 $\pm$ 0.07 & 5.85 $\pm$ 1.18 & 20.31 $\pm$ 4.04 & 1.21 $\pm$ 0.20 \\ 
    + LS & 0.98 $\pm$ 0.03 & 1.08 $\pm$ 0.02 & 8.29 $\pm$ 2.54 & 17.75 $\pm$ 4.23 & 1.73 $\pm$ 0.06 \\ 
    + TS & 0.96 $\pm$ 0.09 & 1.07 $\pm$ 0.07 & 5.93 $\pm$ 2.45 & 15.45 $\pm$ 3.94 & 1.72 $\pm$ 0.15 \\  
    + LS \& TS & 0.95 $\pm$ 0.04 & 1.05 $\pm$ 0.04 & 4.34 $\pm$ 1.29 & 15.58 $\pm$ 1.89 & 1.59 $\pm$ 0.10 \\  \addlinespace 
    \bottomrule
    \end{tabular}
    \label{table:ce25}
\end{table}

\subsection{Implementation Details}
\label{appendix:implementation}

The models used in this work follow the architecture presented by \citet{qiu2020multilevel}, which has been adapted to suit the reduced number of classes considered. The code can be publicly accessed\footnote{\href{https://github.com/ChunpingQiu/benchmark-on-So2SatLCZ42-dataset-a-simple-tour}{\texttt{https://github.com/ChunpingQiu/benchmark-on-So2SatLCZ42-dataset-a-simple-tour}}}. In particular, we used a network depth of 17 (following from the use of four convolutional layers in each block), a width of 16, a dropout rate of 0.2 (at the end of the second and third block) and activated multi-level feature fusion and double-pooling. Class weights did not lead to improved results, hence they were discarded since also the class imbalance differed largely between the training, validation, and testing set. The Nesterov Adam optimizer implementation of Keras \cite{chollet2015keras} was used for training. An early stopping mechanism was installed, which monitored the validation loss with a patience of 20 epochs. Weights were saved after every epoch if and only if the validation loss decreased. 

Label smoothing was performed with a smoothing parameter of 0.1. Temperature scaling was implemented as described by \citet{guo2017calibration} via tuning the scaling parameter on the validation set with respect to minimizing the negative log-likelihood. This minimization was performed using the Adam optimizer implementation of Keras \cite{chollet2015keras} with a learning rate of 0.01 and a maximum of 10k iterations (more iterations did not improve the results significantly). It is worth noting, that temperature scaling scales the logits, but does not change the accuracy of the model, being the reason why we did not include the metric in Table \ref{table:performance}. 

Regarding the hyperparameters of Sen2LCZ, we set a batch size of 64 and an initial learning rate of $2 \times 10^{-3}$ which was gradually reduced by a factor of $0.5$ every five epochs. The ranges of the hyperparameters were as follows: learning rate $\in [2 \times 10^{-1},2 \times 10^{-2},2 \times 10^{-3},2 \times 10^{-4},2 \times 10^{-5}]$, batch size $\in [16,32,64,128]$, smoothing parameter [0.05,0.1,0.2,0.3,0.4,0.5], dropout rate [0.1,0.2,0.3,0.4,0.5], stepsize for learning rate scheduler $\in [2,3,4,5,6,7,8]$, learning rate decay $\in [0.3,0.4,0.5,0.6,0.7]$. For the optimization of the negative log likelihood within the validation data used for steering the temperature, we considered: learning rate $\in [10^{-1},10^{-2},10^{-3},10^{-4},10^{-5}]$, max steps $\in [1k,10k,20k,50k,100k]$. We tried different numbers of predictions for the Monte Carlo Dropout approach, yet more than 20 predictions did not change the derived metrics much. For training, an NVIDIA Tesla P100 GPU was used. Due to the relatively short training time per model (on average < 1 hour), the overall computational cost including hyperparameter grid search was reasonable (ca. 150 - 200 GPU hours). 

\end{document}